\documentclass[11pt]{article}

\usepackage[margin=1in]{geometry}
\usepackage{amsmath}
\usepackage{graphicx}
\usepackage{tikz}
\PassOptionsToPackage{hyphens}{url}
\usepackage{url}
\usepackage{array}
\usepackage{booktabs}
\usepackage{tabularx}
\usepackage{hyperref}
\usetikzlibrary{arrows.meta,positioning,fit,shapes.geometric}

\title{Representation Signatures and Risk-Feedback Alignment in LLM Trading Agents}
\author{Weicheng Xue\\Virginia Tech\\\texttt{weich97@vt.edu}}
\date{}

\begin{document}
\maketitle

\begin{abstract}
We study behavioral alignment and representation dynamics of large language model (LLM) agents in financial decision environments. TradeArena, an auditable trading-agent testbed with risk reports, execution simulation, memory, and replayable trajectories, lets us analyze how rationales, positions, and interventions evolve under market stress. Code and data artifacts are available through the \href{https://github.com/weich97/TreLLM-public}{TradeArena repository}. We find pre-failure signatures: planning embeddings drift from normal centroids, fused plan-risk representations separate normal from pre-drawdown states, and local manifolds exhibit effective-rank contraction. Across 80 rolling failure anchors and eight LLM trajectories, this pattern persists across hash, LSA, Transformer, and white-box hidden-state probes. Stress tests with CoT-free target weights, lexical controls, OHLCV noise, and false audits show that rationale-level contraction can vanish without rationales, while intent-space and fused signatures remain informative. Structured risk feedback can act as an external alignment signal without fine-tuning, but not as a universal performance enhancer: true audit feedback improves calibration for some models, returns for others, and exposes cases where placebo or hidden feedback has higher short-horizon return but weaker alignment diagnostics. A 51-stock intraday experiment reveals a correlation blind spot: LLM rationales justify exposure to coupled assets that the risk layer clips. Finally, a financial-audit task suite shifts comparison from ``which model trades best'' to whether models can audit trajectories, respect execution boundaries, reproduce artifacts, and avoid claim overreach. These results support a research claim, not a profitability claim: auditable risk feedback and representation trajectories reveal when LLM financial reasoning is aligning, drifting, or failing.
\end{abstract}

\section{Introduction}

LLM agents have made it easier to build systems that observe data, call tools, produce rationales, and act in sequential environments. Finance is a natural but high-stakes setting for these systems: an agent may combine market data, news, macro signals, memory, portfolio state, and risk constraints before proposing trades. Yet many evaluations collapse this process into a single return curve or an idealized backtest. Such evaluations can hide whether the agent used future information, whether its risk constraints were enforced, whether execution assumptions were realistic, whether statistical uncertainty is material, or whether another researcher can reconstruct the decision.

This paper studies a scientific question rather than only presenting a tool: when LLM agents make financial decisions under explicit constraints, do their representations and intentions reveal measurable signatures of alignment or failure? Our thesis is that the answer is yes, but only if the evaluation records the full observe-plan-risk-act-reflect lifecycle. A return curve alone cannot show whether a model anticipated a drawdown, hallucinated unsupported context, learned from a risk report, or repeatedly proposed concentrated exposure that an external risk layer had to correct.

The scientific contributions are organized around the decision dynamics rather than the software artifact:
\begin{itemize}
  \item \textbf{Representation signatures of failure}: we show that LLM planning and fused plan-risk representations shift before drawdown troughs and exhibit effective-rank contraction across hash, LSA, BGE-M3 Transformer, CoT-free, and noise-injected diagnostic views.
  \item \textbf{Risk-feedback alignment and over-alignment}: we show that structured risk reports can change subsequent model intent in context, while placebo and contrarian reports can induce excessive conservatism without the same performance benefit.
  \item \textbf{Limits of high-dimensional financial reasoning}: we identify a correlation-blindness failure mode in which an LLM assigns high intent to strongly coupled equity pairs using name-level rationales.
  \item \textbf{Financial-audit agent reliability}: we evaluate frontier models as reviewers of trajectories, risk reports, execution assumptions, reproducibility fields, and claim boundaries, separating audit competence from trading-profit claims.
  \item \textbf{Auditable experimental substrate}: TradeArena is the substrate that makes these claims measurable through replayable trajectories, risk reports, execution events, and representation diagnostics.
\end{itemize}

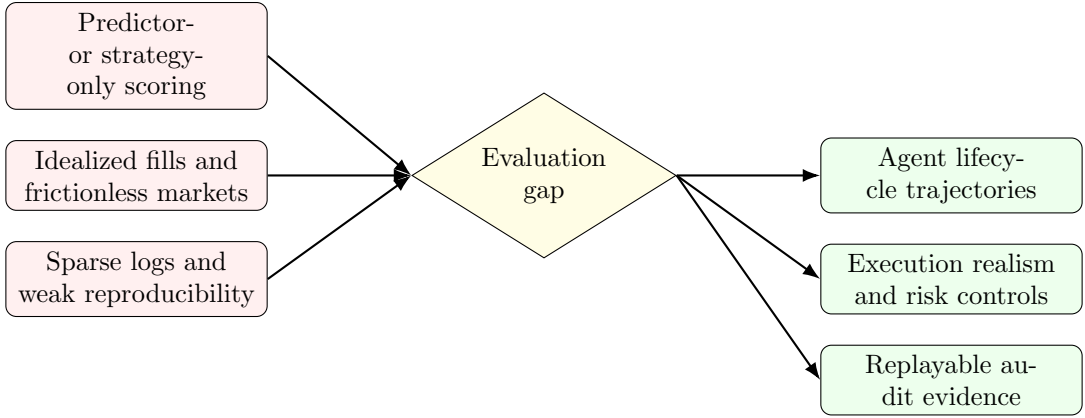
\begin{figure}[t]
  \centering
  \begin{tikzpicture}[font=\small,>=Latex,node distance=1.2cm]
    \tikzstyle{box}=[draw, rounded corners, align=center, minimum height=0.9cm, text width=3.2cm]
    \node[box, fill=red!6] (old1) {Predictor- or strategy-only scoring};
    \node[box, fill=red!6, below=0.4cm of old1] (old2) {Idealized fills and frictionless markets};
    \node[box, fill=red!6, below=0.4cm of old2] (old3) {Sparse logs and weak reproducibility};
    \node[draw, diamond, aspect=1.6, fill=yellow!12, right=1.9cm of old2, align=center] (gap) {Evaluation\\gap};
    \node[box, fill=green!7, right=1.9cm of gap] (new1) {Agent lifecycle trajectories};
    \node[box, fill=green!7, below=0.4cm of new1] (new2) {Execution realism and risk controls};
    \node[box, fill=green!7, below=0.4cm of new2] (new3) {Replayable audit evidence};
    \draw[->, thick] (old1.east) -- (gap.west);
    \draw[->, thick] (old2.east) -- (gap.west);
    \draw[->, thick] (old3.east) -- (gap.west);
    \draw[->, thick] (gap.east) -- (new1.west);
    \draw[->, thick] (gap.east) -- (new2.west);
    \draw[->, thick] (gap.east) -- (new3.west);
  \end{tikzpicture}
  \caption{Motivation. TradeArena turns the evaluation target from a headline return into an accountable trading-decision system.}
  \label{fig:motivation}
\end{figure}

\section{Related Work}

\textbf{Portfolio evaluation and risk.}
Classical portfolio theory formalized the trade-off between expected return and risk \cite{markowitz1952}, while Sharpe-style reward-to-variability performance measures remain common summary statistics for comparing trading strategies \cite{sharpe1966}. TradeArena adopts these familiar metrics but does not treat them as sufficient. Instead, performance is reported alongside drawdown, behavior, execution, risk-audit, and reproducibility measures.

\textbf{Market execution and backtest realism.}
Execution quality can materially change strategy outcomes. Optimal execution work studies the cost and risk of trading over time \cite{almgren2001}, and backtest-overfitting work warns that apparent performance can be an artifact of selection and evaluation design \cite{bailey2014}. TradeArena responds by making execution assumptions explicit: realistic runs include commissions, slippage, latency, liquidity constraints, rejected orders, pending orders, and partial fills, while ideal-execution runs are treated as ablations rather than as the default evidence.

\textbf{LLM agents and tool use.}
Modern language-agent research emphasizes that useful agents often interleave reasoning, tool use, memory, and action. ReAct-style agents combine reasoning traces with actions \cite{react2023}, Toolformer studies language models that call tools \cite{toolformer2023}, and generative-agent work illustrates the importance of memory and state in long-running agent behavior \cite{generativeagents2023}. TradeArena adapts this agent-native perspective to trading: every decision step can include observation, planning, tool outputs, risk review, execution, reflection, and memory updates.

\textbf{Representation geometry and collapse.}
Representation-learning work has shown that contextual embedding spaces can be anisotropic and have strongly non-isotropic geometry \cite{ethayarajh2019}, and neural-collapse analyses study how representations can contract toward lower-dimensional structure under strong training pressure \cite{papyan2020}. Our setting is intentionally different. Classic neural collapse concerns hidden activations in the terminal phase of supervised training, often in the last-layer geometry around class centroids. We do not claim to observe hidden-state collapse inside an LLM, nor do we inspect training dynamics or internal activations. We borrow the geometric language only as an external behavioral diagnostic: whether the observable text, risk, and intent representations of an acting agent become lower-rank before financial failure. Effective rank is therefore used to quantify cognitive narrowing and loss of decision diversity in the agent's external action-and-rationale trace, not as a claim about internal neural states.

\textbf{Cognitive framing.}
Financial decisions are also shaped by behavioral biases under uncertainty \cite{kahneman1979}. We use this perspective only as an interpretive frame, not as a psychological claim about LLM internals. The relevant analogy is that language models may prefer coherent name-level narratives over second-order statistical structure, and may exhibit a form of decision-space narrowing under stress. These ideas motivate our analysis of narrative bias in high-dimensional portfolios and cognitive narrowing in pre-failure representation geometry.

\textbf{Positioning.}
The goal is not to replace financial backtesting engines or to claim live-market profitability. TradeArena is an auditable research substrate for making finance-agent decision dynamics inspectable. Its closest methodological concern is the gap between a headline return curve and the evidence needed to reconstruct, audit, and stress-test each agent decision.

Table~\ref{tab:framework_comparison} summarizes the relationship to existing work. FinRL and FinRL-Meta emphasize reinforcement-learning environments and financial market benchmarks \cite{finrl2020,finrlmeta2022}; Qlib provides an AI-oriented quantitative investment platform with data and model workflows \cite{qlib2020}; FinGPT focuses on financial large language models and data-centric adaptation \cite{fingpt2023}; TradingAgents explores multi-agent LLM trading organizations \cite{tradingagents2024}. TradeArena is complementary: it records the evidence needed to test representation drift, risk-feedback alignment, and execution-sensitive decision failures.

\begin{table}[t]
  \centering
  \caption{Auditable-evidence comparison with related systems. The table compares emphasis rather than live-market performance.}
  \label{tab:framework_comparison}
  \resizebox{\linewidth}{!}{
  \begin{tabular}{lccccc}
    \toprule
    System & Main focus & Agent trace & Risk lifecycle & Execution realism & Raw trajectories \\
    \midrule
    FinRL / FinRL-Meta & RL environments & Partial & Limited & Environment-dependent & Limited \\
    Qlib & Quant research platform & No & Limited & Backtest-oriented & Limited \\
    FinGPT & Financial LLMs & Partial & No & No & No \\
    TradingAgents & Multi-agent LLM trading & Yes & Partial & Limited & Partial \\
    TradeArena & Auditable agent benchmark & Yes & Yes & Yes & Yes \\
    \bottomrule
  \end{tabular}}
\end{table}

\section{Design Goals}

TradeArena is organized around six design goals.
\textbf{Modularity}: data providers, analysts, strategies, risk gates, execution simulators, memory stores, and evaluators use narrow interfaces.
\textbf{Reproducibility}: every run records observations, signals, decisions, risk reports, orders, fills, portfolio state, memory events, and metrics.
\textbf{Execution realism}: the realistic simulator accounts for commissions, slippage, latency, participation limits, rejected orders, pending orders, and partial fills.
\textbf{Risk awareness}: decisions can be routed through structured pre-trade, in-trade, and post-trade risk reports.
\textbf{Auditability}: trajectories include the evidence needed to explain why a trade happened and what happened after submission.
\textbf{Agent-native evaluation}: each step records the observe-plan-tool-risk-act-reflect lifecycle expected from modern LLM agent systems.

\section{System Architecture}

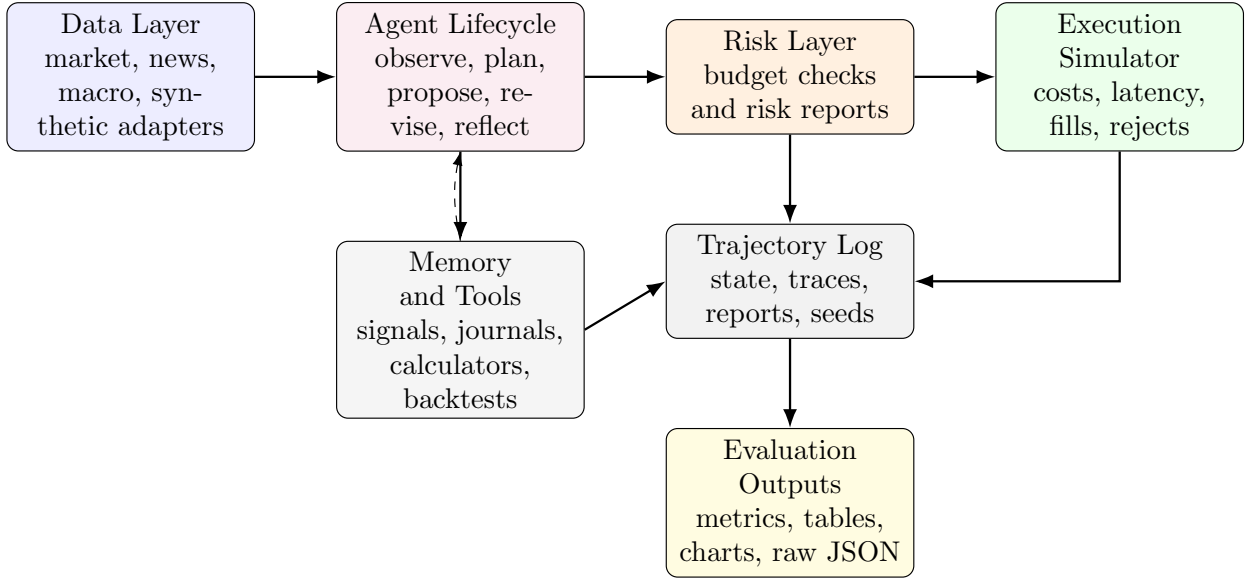
\begin{figure}[t]
  \centering
  \resizebox{\linewidth}{!}{
  \begin{tikzpicture}[font=\small,>=Latex,node distance=1.1cm]
    \tikzstyle{layer}=[draw, rounded corners, align=center, minimum height=1.0cm, text width=2.8cm]
    \node[layer, fill=blue!7] (data) {Data Layer\\market, news, macro, synthetic adapters};
    \node[layer, fill=purple!7, right=1.0cm of data] (agent) {Agent Lifecycle\\observe, plan, propose, revise, reflect};
    \node[layer, fill=orange!12, right=1.0cm of agent] (risk) {Risk Layer\\budget checks and risk reports};
    \node[layer, fill=green!8, right=1.0cm of risk] (exec) {Execution Simulator\\costs, latency, fills, rejects};
    \node[layer, fill=gray!8, below=1.1cm of agent] (memory) {Memory and Tools\\signals, journals, calculators, backtests};
    \node[layer, fill=gray!10, below=1.1cm of risk] (traj) {Trajectory Log\\state, traces, reports, seeds};
    \node[layer, fill=yellow!14, below=1.1cm of traj] (eval) {Evaluation Outputs\\metrics, tables, charts, raw JSON};
    \draw[->, thick] (data) -- (agent);
    \draw[->, thick] (agent) -- (risk);
    \draw[->, thick] (risk) -- (exec);
    \draw[->, thick] (exec.south) |- (traj.east);
    \draw[->, thick] (risk.south) -- (traj.north);
    \draw[->, thick] (agent.south) -- (memory.north);
    \draw[->, thick] (memory.east) -- (traj.west);
    \draw[->, thick] (traj) -- (eval);
    \draw[->, dashed] (memory.north) to[bend left=12] (agent.south);
  \end{tikzpicture}}
  \caption{TradeArena architecture. Components are replaceable, but all routes converge into replayable trajectories and evaluation outputs.}
  \label{fig:architecture}
\end{figure}

The core package exposes plugin-like interfaces for data, analysis, strategy, risk, execution, memory, and evaluation. This separation is intentional. Trading-agent research often couples data access, prompting, portfolio construction, execution, and evaluation into one pipeline, making it difficult to identify whether an observed gain comes from better signals, looser risk constraints, unrealistic fills, or evaluation leakage. TradeArena instead isolates these concerns so that one component can be replaced while the rest of the benchmark remains fixed.

The architecture in Figure~\ref{fig:architecture} reflects four design choices.
\textbf{First}, observations are separated from decisions: the data layer provides timestamped market and contextual inputs, while the agent lifecycle records how those inputs are converted into intent.
\textbf{Second}, risk is a first-class stage rather than an after-the-fact metric: the risk layer can clip, block, or annotate decisions before they reach the execution simulator.
\textbf{Third}, execution is modeled as an environment response: orders may be delayed, partially filled, rejected, or filled with slippage and commissions.
\textbf{Fourth}, every stage writes to a shared trajectory, which becomes the evidence base for metrics, charts, and per-step audit inspection. The reference benchmark uses deterministic synthetic market data so that these mechanics can be reproduced without external data access; real data adapters and model-backed agents can be added without changing the logging contract.

\section{Experimental Setup}

\subsection{Research Questions}

The experiments are organized around five research questions rather than a long list of narrowly scoped claims:
\begin{enumerate}
  \item \textbf{Core systems}: How do risk gates, execution assumptions, stress conditions, and component ablations change benchmark behavior?
  \item \textbf{Robustness}: Are the main conclusions stable across random seeds, heterogeneous synthetic markets, historical rolling windows, and intraday execution stress?
  \item \textbf{Agent dynamics}: Do memory, risk feedback, and cross-model reflection alter intended exposure after the risk layer intervenes?
  \item \textbf{Representation and trust}: Can plan/reflect embeddings and hallucination proxies expose decision drift and unsupported reasoning?
  \item \textbf{Scalability}: Does the benchmark remain informative on a 50+ stock, 1-hour portfolio task with nontrivial cross-asset correlation?
\end{enumerate}

\subsection{Benchmark Protocol and Baselines}

The experiment suite uses three synthetic assets, 120 decision periods, and seeds 3, 7, and 11 for the full trajectory-generating benchmark. All cases share the same initial capital, evaluation code, logging schema, and trajectory writer. The expanded suite combines synthetic controls, historical-market cases, direct-provider LLM sanity checks, and a cached Poe-mediated frontier model matrix. To strengthen statistical reliability without bloating the study with hundreds of raw trajectories, the evaluation also runs a 30-seed robustness sweep for the core synthetic cases, a 120-market heterogeneous synthetic stress test, and a four-window rolling historical validation. These robustness sweeps report aggregate tables with confidence intervals, paired differences, win rates, and p-values rather than per-step JSON trajectories. Table~\ref{tab:experiment_matrix} lists the experiment families.

\begin{table}[t]
  \centering
  \caption{Expanded experimental matrix. Full synthetic trajectories use seeds 3, 7, and 11; statistical robustness uses seeds 1--30; historical robustness uses rolling two-year windows.}
  \label{tab:experiment_matrix}
  \scriptsize
  \begin{tabularx}{\linewidth}{>{\raggedright\arraybackslash}p{0.25\linewidth}>{\raggedright\arraybackslash}p{0.10\linewidth}>{\raggedright\arraybackslash}X}
    \toprule
    Family & Cases & Purpose \\
    \midrule
    Core baselines & 4 & Proposed agent vs buy-and-hold, ideal execution, no risk gate \\
    Execution sweep & 3 & Ideal, realistic, and constrained-latency execution \\
    Risk sensitivity & 3 & Strict, default, and loose max-position policies \\
    Analyst ablation & 3 & Momentum-only, macro/news-only, and full analyst stack \\
    Memory ablation & 2 & Memory-blind vs memory-aware signal strategy \\
    Historical market data & 4 & GSPC, BTC-USD, ETH-USD, 2021--2026 weekly decisions \\
    LLM historical sanity checks & 5 & Direct-provider model analysts versus deterministic recent-window baseline \\
    Frontier model matrix & 5 & Cached Poe-mediated GPT-5.5, Gemini 3.1 Pro, Kimi K2.5, GLM-5, and Claude Opus 4.7 risk-aware agents \\
    Frontier feedback matrix & 15 & Cached Poe-mediated true, placebo, and hidden risk-feedback ablations \\
    CoT-free frontier ablation & 6 & GPT-5.5, Gemini, and Claude with rationales removed and JSON target weights only \\
    Contrarian audit probe & 6 & True versus false severe risk reports for frontier trust calibration \\
    Statistical seed sweep & 5 & 30-seed confidence intervals and paired differences \\
    Synthetic market stress & 240 pairs & 120 heterogeneous volatility, tail, jump, and trend markets with p-values \\
    Rolling-window validation & 12 & Four two-year windows over the historical market data \\
    LLM representation drift & 8 & Plan/reflect embedding shifts and rolling failure anchors across provider-specific and frontier LLMs \\
    Lexical collapse controls & 6 & Type-token ratio and token entropy around the same rolling failure anchors \\
    Noise-injection robustness & 8 & Pre-failure signatures under 0, 5, 10, and 20\% OHLCV perturbations across three frontier models \\
    White-box hidden-state probe & 4 & Qwen2.5-0.5B last-layer hidden states versus LSA decision-text embeddings \\
    Hallucination-risk audit & 4 + 50 & Unsupported-claim proxy plus blind human-calibration annotation template \\
    Memory learning curves & 2 & Risk-gate frequency and calibration over 52 LLM decision steps \\
    Risk-feedback ablation & 3 & Direct-provider true, placebo, and hidden structured risk reports \\
    Intraday 50-stock portfolio & 7 & 1-hour Yahoo Finance bars, Markowitz baseline, cached Poe-mediated 40-step LLM probes, correlation complexity, and execution stress \\
    Financial-audit skill tasks & 500 calls & Poe-mediated standard, challenge, follow-up, and adversarial artifact-review tasks \\
    Stress tests & 4 & High cost, low liquidity, latency, fragile microstructure \\
    \bottomrule
  \end{tabularx}
\end{table}

The core comparison includes one proposed agent and three baselines or ablations:
\begin{enumerate}
  \item \textbf{Risk-aware realistic agent}: the main TradeArena reference agent, using momentum and macro/news signals, a max-position risk gate, and realistic execution.
  \item \textbf{Buy-and-hold realistic baseline}: an equal-weight buy-and-hold portfolio under the same realistic execution simulator. This is the primary baseline because it tests whether the agent adds value beyond passive exposure to the synthetic assets.
  \item \textbf{Ideal-execution ablation}: the same signal-driven agent under idealized execution. This isolates the effect of optimistic fills and missing market frictions.
  \item \textbf{No-risk ablation}: the same signal-driven agent with the risk gate disabled. This isolates the effect of the risk layer on fills, rejections, violations, and return.
\end{enumerate}

The suite also runs four stress tests: high transaction costs, low liquidity, high latency, and fragile microstructure with high impact and constrained participation. Each run writes CSV tables, SVG charts, raw per-case trajectory JSON files, and a summary JSON file.

The extended ablations are implemented as first-class benchmark cases rather than post-hoc calculations. The execution sweep changes the simulator while holding the agent fixed. The risk-sensitivity sweep changes maximum position and turnover limits while holding signals and execution fixed. The analyst ablation removes one source of signals at a time. The memory ablation compares the default signal-weighted strategy with a memory-aware overlay that reduces exposure after recent drawdowns, rejected orders, or risk events and modestly increases exposure after stable positive recent performance.

\subsection{Historical Market Data}

To address the limitation of purely synthetic evaluation, TradeArena includes a real-data experiment using Yahoo Finance daily OHLCV data from May 2021 to May 2026. The universe contains the S\&P 500 index proxy \texttt{GSPC}, Bitcoin \texttt{BTC-USD}, and Ethereum \texttt{ETH-USD}. The benchmark uses weekly decision frequency over the downloaded daily history, preserving the same agent, risk, execution, trajectory, and logging interfaces used in the synthetic benchmark. We do not claim that one-hour or sub-hour execution latency dominates weekly returns. Instead, the low-frequency experiment uses the execution simulator as a conservative audit layer: even weekly portfolio changes can be affected by liquidity participation limits, partial fills, rejected orders, and large gaps during volatile or liquidity-constrained periods. The stronger test of execution realism is therefore the separate 1-hour intraday experiment described below. The historical experiment includes four deterministic cases: risk-aware realistic, buy-and-hold realistic, ideal execution, and no-risk realistic. It also includes shorter 52-week LLM decision experiments that replace the deterministic analyst stack with direct-provider model analysts while holding the same market universe, risk gate, execution simulator, and risk-feedback prompt. The primary cross-model comparison is a cached Poe-mediated frontier matrix evaluating \texttt{gpt-5.5}, \texttt{gemini-3.1-pro}, \texttt{kimi-k2.5}, \texttt{glm-5}, and \texttt{claude-opus-4.7} under the same risk-aware protocol and under true, placebo, and hidden feedback conditions. These rows replay local prompt/response cache entries keyed by provider model names; the public repository tracks derived tables and redacted cache manifests rather than raw provider text. The released adapter can also generate new rows through Poe's OpenAI-compatible endpoint when \texttt{POE\_API\_KEY} is set, while \texttt{OPENAI\_API\_KEY} is not used for these experiments. Because commercial model aliases are version-sensitive experimental entities rather than archival papers, we report exact provider strings, platform routes, and access dates in the artifact metadata rather than treating those pages as scholarly references. The LLM prompt contains short-term feedback from the most recent decisions and a long-term 52-step risk memory summarizing clipped decisions, blocked decisions, risk violations, rejected orders, slippage, and recent failure examples. For robustness, the study additionally evaluates four rolling windows: 2021--2023, 2022--2024, 2023--2025, and 2024--2026. This experiment is not intended to be a production trading study; its role is to test whether the framework exposes realistic behavior on volatile historical assets and whether model-backed reasoning can be audited under the same logging contract.

\subsection{Intraday 50-Stock Portfolio Data}

To increase market complexity, TradeArena also includes a 1-hour intraday experiment over 51 liquid U.S. equities spanning technology, financials, healthcare, consumer, energy, industrials, and utilities. The data are downloaded with the Yahoo Finance chart endpoint using \texttt{range=60d} and \texttt{interval=1h}, matching the public yfinance interval support and intraday lookback limit \cite{yfinance_docs}. The repository records a manifest with ticker list, interval, range, and download timestamp. The downloaded panel contains 414 aligned 1-hour bars after timestamp intersection. The experiment computes a full pairwise return-correlation summary over the 51-stock panel and then evaluates the most recent 40 hourly decisions, roughly one trading week, for buy-and-hold, deterministic risk-aware allocation, a rolling Markowitz minimum-variance optimizer, low-liquidity stress, latency stress, and two cached Poe-mediated frontier LLM probes: GPT-5.5 and Gemini 3.1 Pro. Each LLM step requires a 51-symbol structured JSON allocation and is recorded through the same cache, risk gate, execution simulator, and trajectory schema as the other LLM experiments.

The intraday task is a cognitive and mechanistic probe, not a claim about statistically significant one-week trading alpha. Its purpose is to test whether an autoregressive agent can transform a dense 51-asset prompt into portfolio intent while respecting second-order dependence. The 40-hour horizon supplies 2,040 per-symbol LLM allocation decisions and 51-by-51 covariance context, which is sufficient for diagnosing attention allocation, risk-gate pressure, and correlated-pair over-weighting, but not sufficient for estimating long-run economic performance. We therefore interpret the intraday results as evidence about high-dimensional portfolio reasoning under realistic execution constraints rather than as a conventional backtest.

\section{Metrics}

TradeArena reports metrics across six categories. Performance metrics include total return, Sharpe ratio, volatility, maximum drawdown, and final equity. Behavioral metrics include order count, fill count, turnover events, and hold ratio. Execution-realism metrics include fill rate, partial-fill rate, rejected orders, pending orders, commissions, slippage cost, latency, and fill ratio. Risk-audit metrics include risk report coverage, blocked or clipped decisions, failed checks, warning checks, and violations. Reproducibility metrics include trajectory coverage, agent trace coverage, and risk lifecycle coverage. Audit-agent metrics score whether a model can inspect a trajectory, identify risk edits and execution frictions, distinguish stress-only evidence from calibrated execution evidence, classify engineering/benchmark/scientific claims, report commit-hash-command provenance, and propose narrow plugins with deterministic tests. These audit-agent scores are not portfolio-return metrics; they measure whether a model can review the evidence that makes a financial-agent benchmark credible.

For the LLM interpretability experiment, TradeArena constructs two reproducible embedding views from plan and reflection records. The primary view uses a fixed 64-dimensional hashing embedding, which guarantees deterministic replay without another model API. As a robustness check, we also compute a corpus-level latent semantic analysis (LSA) embedding from TF-IDF plan documents; unlike the hash view, LSA uses low-rank co-occurrence structure and is a standard semantic embedding baseline. The plan text concatenates LLM rationales, risk notes, and target-weight decisions. The reflection text concatenates post-trade attribution, execution outcomes, rejected or pending order counts, slippage, and risk-violation summaries. A fused representation appends structured market and risk features to the text representation. We label the four steps before a drawdown anchor as \emph{pre-drawdown}, the anchor and following steps as \emph{drawdown}, and all other steps as \emph{normal}. We then measure cosine distances between phase centroids and a nearest-centroid balanced accuracy for distinguishing pre-drawdown from normal states. To move beyond centroid distances, we also compute representation-manifold diagnostics: path length, local step distance, pre-to-normal velocity ratio, effective rank of local embedding neighborhoods, and a nearest-neighbor phase-purity score. The main table reports maximum-drawdown anchors, while the robustness table uses 80 rolling failure anchors across eight LLM trajectories.

Two additional probes test whether the pre-failure signature is merely a language artifact or a market-noise artifact. The CoT-free ablation asks GPT-5.5, Gemini 3.1 Pro, and Claude Opus 4.7 to return only JSON target weights, with no rationale, reflection, or risk note. We then compare language-plan geometry with intent-weight geometry. A lexical-control table computes type-token ratio and token entropy around the same rolling failure anchors. This separates semantic contraction in embedding space from trivial text degeneration such as repeating a few risk words. The noise-injection probe perturbs the historical OHLCV price stream with deterministic Gaussian shocks at \(\epsilon \in \{0.05,0.10,0.20\}\) and recomputes fused market-plan diagnostics around the same rolling failure anchors. A stable signature under noise suggests that the diagnostic is not simply rediscovering raw price volatility.

For the memory-learning experiment, TradeArena tracks two longitudinal diagnostics. The first is risk-gate rate: whether a step triggered clipping, blocking, or a risk violation. The second is decision calibration: the agreement between intended target weights and risk-approved target weights. Calibration gap is the sum of absolute target-weight changes imposed by the risk layer; calibration score is \(1-\min(1,\text{gap}/\max(1,\text{intended exposure}))\). A learning effect is indicated by a falling risk-gate rate and a rising calibration score as the agent accumulates memory.

For the risk-feedback ablation, model analysts are evaluated under three conditions on the same 52-week historical market task. In the true-feedback condition, the prompt includes recent clipped counts, blocked counts, risk violations, rejected orders, pending orders, slippage, equity, and a long-term risk-memory summary. In the hidden condition, the market data and portfolio state are unchanged but these structured reports are removed. In the placebo condition, the prompt retains the same structured fields but replaces the audit values with deterministic counterfactual risk reports. We compare intended absolute exposure, the calibration gap between intended and risk-approved weights, late-window behavior, and intent drift, defined as late-window intended exposure minus early-window intended exposure. This tests whether risk reports act as an external supervision signal for the model's decision intent rather than merely as a downstream filter or a generic conservative prompt. The main version repeats this true/placebo/hidden protocol across the cached Poe-mediated frontier matrix; the direct-provider run is retained as a sanity check.

The contrarian-audit probe is a stronger trust-calibration test. When the realized trajectory is benign, the prompt is injected with a severe but false structured risk report claiming repeated clipping, blocking, risk violations, rejected orders, and large slippage. If the model reacts with a conservative intent shift despite conflicting portfolio evidence, then the feedback channel is not merely an information channel; it is also an attack surface for miscalibrated external supervision.

For the heterogeneous synthetic stress test, TradeArena generates 120 parallel markets with different volatility states, tail states, jump processes, and trend signs. Volatility states range from calm to crisis; tail states include Gaussian shocks, Student-\(t\) shocks, jump shocks, and Student-\(t\) plus jumps. Each market is evaluated with the same paired cases, so the statistical unit is the market rather than an individual time step. We report paired mean differences, 95\% confidence intervals, win rates, and two-sided paired-test p-values. Because the implementation is standard-library only, p-values use the normal approximation to the paired \(t\)-statistic; they should be read as large-sample significance diagnostics rather than exact small-sample tests.

For hallucination-risk analysis, TradeArena constructs an audit-grounded factual-violation proxy rather than a broad open-domain semantic hallucination detector. A step receives proxy mass when the LLM rationale makes a claim that can be mechanically audited against the recorded prompt and trajectory: unsupported external context not present in the prompt, such as news, earnings, regulation, macro, on-chain evidence, support, resistance, or breakouts; a directional score that contradicts the current OHLCV intraperiod move; high confidence despite weak price evidence; or a claim that no prior risk exists after earlier risk or execution failures have already appeared in memory. The proxy is then correlated with risk-gate triggers, risk violations, calibration gaps, and rejected orders using the same recorded trajectory fields. To make optional human calibration reproducible rather than anecdotal, the suite also exports a blind 50-step annotation sample with rationale excerpts, proxy labels, and empty annotator fields; when two human labels are supplied, the code reports inter-annotator Cohen's \(\kappa\) and proxy-versus-adjudicated IoU.

\section{Results}

\subsection{Core Baseline Comparison}

Table~\ref{tab:core} summarizes the mean core results across the three seeds. Against the buy-and-hold baseline, the risk-aware realistic agent increases mean return from 0.257 to 0.423 and improves Sharpe from 2.913 to 7.696. It also reduces maximum drawdown from -0.261 to -0.025. This comparison is important because both cases use the same realistic execution model; the difference is therefore not an artifact of granting the active agent easier fills.

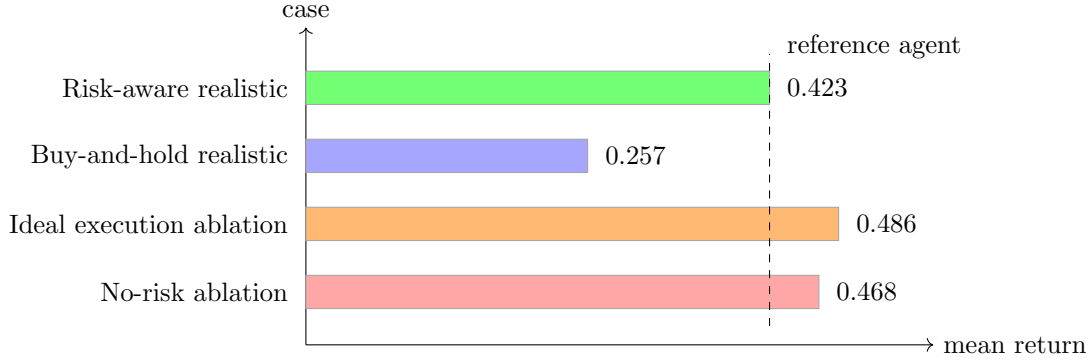
\begin{figure}[t]
  \centering
  \begin{tikzpicture}[font=\small]
    \draw[->] (0,0) -- (8.3,0) node[right] {mean return};
    \draw[->] (0,0) -- (0,4.2) node[above] {case};
    \foreach \y/\name/\val/\color in {
      3.4/Risk-aware realistic/0.423/green!55,
      2.5/Buy-and-hold realistic/0.257/blue!35,
      1.6/Ideal execution ablation/0.486/orange!55,
      0.7/No-risk ablation/0.468/red!35
    } {
      \node[anchor=east] at (-0.1,\y) {\name};
      \draw[fill=\color, draw=black!35] (0,\y-0.22) rectangle ({\val*14.5},\y+0.22);
      \node[anchor=west] at ({\val*14.5+0.1},\y) {\val};
    }
    \draw[dashed] ({0.423*14.5},0.25) -- ({0.423*14.5},3.85);
    \node[anchor=west] at ({0.423*14.5+0.1},3.95) {reference agent};
  \end{tikzpicture}
  \caption{Mean return comparison across the core cases. The ideal-execution row is an ablation, not the default benchmark setting.}
  \label{fig:return_comparison}
\end{figure}

\begin{table}[t]
  \centering
  \caption{Core benchmark means over seeds 3, 7, and 11.}
  \label{tab:core}
  \begin{tabular}{lrrrrrr}
    \toprule
    Case & Return & Sharpe & Vol. & Max DD & Fill & Slippage \\
    \midrule
    Risk-aware realistic & 0.423 & 7.696 & 0.098 & -0.025 & 0.876 & 4306.3 \\
    Buy-and-hold realistic & 0.257 & 2.913 & 0.172 & -0.261 & 0.895 & 769.5 \\
    Ideal execution ablation & 0.486 & 8.878 & 0.095 & -0.019 & 0.965 & 271.0 \\
    No-risk ablation & 0.468 & 7.949 & 0.103 & -0.029 & 0.673 & 4337.4 \\
    \bottomrule
  \end{tabular}
\end{table}

\subsection{Statistical Robustness Across 30 Seeds}

Table~\ref{tab:statistical_significance} reports a larger 30-seed sweep for the core synthetic comparisons. The risk-aware realistic agent has mean return 0.406 with a 95\% confidence interval of [0.385, 0.426]. The buy-and-hold baseline has mean return 0.269 with interval [0.243, 0.295]. The paired return difference between the risk-aware agent and buy-and-hold is 0.137 with interval [0.117, 0.156], which does not cross zero. The same pattern holds for maximum drawdown: the risk-aware agent improves drawdown by 0.234 relative to buy-and-hold, with paired interval [0.228, 0.240].

The 30-seed sweep also clarifies the ablations. Ideal execution has higher return than buy-and-hold, but its advantage is accompanied by a much higher fill rate and much lower slippage cost, confirming that favorable execution assumptions materially change outcomes. The no-risk ablation has the highest mean return in this synthetic setting, 0.465, but its mean fill rate is only 0.669, compared with 0.888 for the risk-aware agent. Under the latency stress case, the paired return advantage over buy-and-hold shrinks to 0.026 with a lower confidence bound close to zero, while fill rate falls sharply. Thus the confidence intervals strengthen the paper's central claim: performance differences should be interpreted jointly with risk and execution diagnostics.

\begin{table}[t]
  \centering
  \caption{Statistical robustness over 30 synthetic seeds. CI denotes a two-sided 95\% confidence interval; paired differences are against buy-and-hold.}
  \label{tab:statistical_significance}
  \resizebox{\linewidth}{!}{
  \begin{tabular}{lrrr}
    \toprule
    Case & Return mean [95\% CI] & Max DD mean [95\% CI] & Return paired diff [95\% CI] \\
    \midrule
    Risk-aware realistic & 0.406 [0.385, 0.426] & -0.025 [-0.028, -0.022] & 0.137 [0.117, 0.156] \\
    Buy-and-hold realistic & 0.269 [0.243, 0.295] & -0.259 [-0.265, -0.253] & -- \\
    Ideal execution & 0.458 [0.438, 0.479] & -0.018 [-0.019, -0.016] & 0.190 [0.171, 0.209] \\
    No-risk realistic & 0.465 [0.440, 0.491] & -0.030 [-0.034, -0.027] & 0.197 [0.173, 0.220] \\
    Latency stress & 0.295 [0.268, 0.322] & -0.056 [-0.061, -0.050] & 0.026 [0.001, 0.052] \\
    \bottomrule
  \end{tabular}
  }
\end{table}

\subsection{Heterogeneous Synthetic Market Stress Test}

Table~\ref{tab:synthetic_market_stress} reports the 120-market synthetic stress test. Across all heterogeneous markets, the risk-aware agent outperforms buy-and-hold by 0.153 total return on average, with a 95\% confidence interval of [0.122, 0.184], win rate 0.842, and \(p<10^{-12}\). It also improves maximum drawdown by 0.243 on average and reduces rejected orders by 11.6 orders per market. These effects remain significant under Gaussian, Student-\(t\), jump, and Student-\(t\)-plus-jump tails.

The result is not uniformly favorable on every metric. The risk-aware agent has a slightly lower fill rate than buy-and-hold because it actively trades and therefore creates more execution opportunities. In crisis-volatility markets, the return advantage over buy-and-hold is not statistically significant (mean 0.049, \(p=0.288\)), but the drawdown improvement remains large and significant. Compared with the no-risk ablation, the risk-aware agent has lower return on average (-0.045, \(p<10^{-9}\)) but materially better drawdown, much higher fill rate, and 38.2 fewer rejected orders per market. This is exactly the trade-off the framework is designed to expose: risk controls may reduce raw return in some synthetic regimes, but they improve operating stability and execution validity.

\begin{table}[t]
  \centering
  \caption{Heterogeneous synthetic stress test over 120 parallel markets. Values are paired differences. Positive return, drawdown, and fill-rate differences favor the risk-aware agent; negative rejected-order differences indicate fewer rejected orders.}
  \label{tab:synthetic_market_stress}
  \resizebox{\linewidth}{!}{
  \begin{tabular}{lllrrrr}
    \toprule
    Comparison & Regime & Metric & \(n\) & Mean diff [95\% CI] & \(p\) & Win rate \\
    \midrule
    Risk-aware vs buy-hold & All & Return & 120 & 0.153 [0.122, 0.184] & $<10^{-12}$ & 0.842 \\
    Risk-aware vs buy-hold & All & Max DD & 120 & 0.243 [0.228, 0.258] & $<10^{-12}$ & 0.983 \\
    Risk-aware vs buy-hold & Crisis vol. & Return & 30 & 0.049 [-0.046, 0.144] & 0.288 & 0.600 \\
    Risk-aware vs buy-hold & Crisis vol. & Max DD & 30 & 0.199 [0.150, 0.247] & $<10^{-12}$ & 0.933 \\
    Risk-aware vs no-risk & All & Return & 120 & -0.045 [-0.059, -0.030] & $<10^{-9}$ & 0.225 \\
    Risk-aware vs no-risk & All & Max DD & 120 & 0.015 [0.011, 0.019] & $<10^{-11}$ & 0.800 \\
    Risk-aware vs no-risk & All & Fill rate & 120 & 0.200 [0.188, 0.212] & $<10^{-12}$ & 1.000 \\
    Risk-aware vs no-risk & All & Rejected orders & 120 & -38.150 [-40.473, -35.827] & $<10^{-12}$ & 0.000 \\
    \bottomrule
  \end{tabular}
  }
\end{table}

\subsection{Execution Realism Ablation}

The ideal-execution ablation is the strongest performing core case, with mean return 0.486 and Sharpe 8.878. However, this improvement comes with a very different execution profile: fill rate increases to 0.965 and slippage cost drops to 271.0, compared with fill rate 0.876 and slippage cost 4306.3 for realistic execution. This gap is precisely why TradeArena treats ideal execution as an ablation. A return-only evaluation would make the idealized result look uniformly better; the execution metrics reveal that part of the improvement is due to more favorable environment assumptions.

Table~\ref{tab:execution_sweep} extends this comparison with a constrained-latency execution setting. Moving from realistic default execution to constrained latency reduces return from 0.423 to 0.330, reduces fill rate from 0.876 to 0.722, and increases slippage cost from 4306.3 to 6621.1. Thus the benchmark captures not only ideal-versus-realistic differences but also graded degradation under more adverse execution.

There is an important frequency caveat. In the weekly historical experiments, execution realism should be interpreted as an audit mechanism for large rebalance orders, liquidity limits, and stress-period partial fills rather than as a claim about millisecond execution alpha. For this reason, the paper's strongest execution-realism evidence comes from the synthetic stress tests and the 1-hour 51-stock intraday experiment, where latency, participation limits, and market impact operate at a frequency closer to the decision interval.

\begin{table}[t]
  \centering
  \caption{Execution sweep means over seeds 3, 7, and 11.}
  \label{tab:execution_sweep}
  \begin{tabular}{lrrrrr}
    \toprule
    Case & Return & Sharpe & Max DD & Fill & Slippage \\
    \midrule
    Ideal & 0.486 & 8.878 & -0.019 & 0.965 & 271.0 \\
    Realistic default & 0.423 & 7.696 & -0.025 & 0.876 & 4306.3 \\
    Constrained latency & 0.330 & 5.812 & -0.048 & 0.722 & 6621.1 \\
    \bottomrule
  \end{tabular}
\end{table}

\subsection{Risk-Gate Ablation}

Disabling the risk gate produces mean return 0.468, higher than the risk-aware realistic agent in this synthetic benchmark. The same row also shows a lower fill rate of 0.673 and slippage cost 4337.4. In the raw metrics, the no-risk ablation averages 62.3 rejected orders, compared with 22.0 for the risk-aware agent. The interpretation is therefore not simply that risk control increases or decreases return; rather, the risk layer changes the operating regime of the strategy. It clips position targets and reduces rejected orders, while preserving a complete risk-lifecycle audit trail. This is a central example of the paper's thesis: a benchmark should report how performance was achieved, not only the final return.

\subsection{Risk Sensitivity}

Table~\ref{tab:risk_sensitivity} varies the maximum position and turnover policy. A strict 20\% position policy has the lowest return, 0.243, but also the lowest volatility and smallest drawdown. A loose 50\% policy increases return to 0.459 and Sharpe to 7.876, but fill rate falls and slippage rises. This sweep is useful because it turns the risk layer into an experimental variable. Instead of treating ``risk control'' as a binary on/off choice, TradeArena can measure the trade-off between risk constraints, execution quality, and performance.

\begin{table}[t]
  \centering
  \caption{Risk-sensitivity means over seeds 3, 7, and 11.}
  \label{tab:risk_sensitivity}
  \begin{tabular}{lrrrrrr}
    \toprule
    Case & Return & Sharpe & Vol. & Max DD & Fill & Clipped \\
    \midrule
    Strict 20\% & 0.243 & 7.138 & 0.065 & -0.018 & 0.926 & 128.3 \\
    Default 35\% & 0.423 & 7.696 & 0.098 & -0.025 & 0.876 & 143.0 \\
    Loose 50\% & 0.459 & 7.876 & 0.102 & -0.026 & 0.854 & 133.3 \\
    \bottomrule
  \end{tabular}
\end{table}

\subsection{Analyst and Memory Ablations}

Table~\ref{tab:agent_ablations} studies two agent-internal questions. First, the analyst ablation shows that momentum is the dominant synthetic signal in the current benchmark: momentum-only return is 0.426, close to the full stack, while macro/news-only return is 0.053. This does not imply macro/news signals are unimportant in real markets; rather, it validates that the benchmark can localize which component explains performance in a controlled environment. Second, the memory-aware strategy modestly improves return, Sharpe, drawdown, and fill rate relative to the memory-blind signal strategy. The effect is intentionally small because the memory overlay is conservative; its role is to demonstrate that memory can be evaluated as an auditable component rather than hidden inside an agent prompt.

\begin{table}[t]
  \centering
  \caption{Analyst and memory ablation means over seeds 3, 7, and 11.}
  \label{tab:agent_ablations}
  \begin{tabular}{lrrrrr}
    \toprule
    Case & Return & Sharpe & Max DD & Fill & Rejected \\
    \midrule
    Momentum only & 0.426 & 7.669 & -0.022 & 0.878 & 18.7 \\
    Macro/news only & 0.053 & 1.904 & -0.064 & 0.911 & 13.0 \\
    Full analyst stack & 0.423 & 7.696 & -0.025 & 0.876 & 22.0 \\
    Memory-blind strategy & 0.423 & 7.696 & -0.025 & 0.876 & 22.0 \\
    Memory-aware strategy & 0.427 & 7.878 & -0.022 & 0.890 & 19.3 \\
    \bottomrule
  \end{tabular}
\end{table}

\subsection{Stress Tests}

Table~\ref{tab:stress} summarizes stress-test means. Latency and fragile microstructure reduce return and Sharpe relative to the standard realistic setting, while fragile microstructure produces the highest slippage cost. High latency is especially damaging to fill quality: fill rate falls to 0.611. Fragile microstructure produces the largest cost burden, with slippage 7771.8. These results are not intended to prove a profitable trading strategy. Instead, they demonstrate that the benchmark can surface execution-sensitive failure modes and produce audit trails for subsequent inspection.

\begin{table}[t]
  \centering
  \caption{Stress-test means over seeds 3, 7, and 11.}
  \label{tab:stress}
  \begin{tabular}{lrrrrrr}
    \toprule
    Case & Return & Sharpe & Vol. & Max DD & Fill & Slippage \\
    \midrule
    High cost & 0.383 & 7.079 & 0.098 & -0.031 & 0.871 & 6330.6 \\
    Low liquidity & 0.417 & 7.614 & 0.098 & -0.026 & 0.884 & 4775.6 \\
    Latency & 0.276 & 4.993 & 0.105 & -0.052 & 0.611 & 5437.9 \\
    Fragile microstructure & 0.307 & 5.467 & 0.104 & -0.052 & 0.722 & 7771.8 \\
    \bottomrule
  \end{tabular}
\end{table}

\subsection{Historical Market Results}

Table~\ref{tab:real_market} reports the five-year historical backtest on GSPC, BTC-USD, and ETH-USD. The real-data experiment is deliberately less flattering than the synthetic benchmark. The risk-aware realistic agent returns 0.136 with Sharpe 0.512, while buy-and-hold returns 0.143 with Sharpe 0.629. However, the risk-aware agent reduces maximum drawdown from -0.626 to -0.366. This is a useful result for the paper: TradeArena does not simply produce high synthetic returns; on real volatile assets it reveals a return/drawdown trade-off and a large gap between realistic and idealized execution.

The ideal-execution historical case returns 0.965 with Sharpe 1.544 and no rejected orders, while the realistic risk-aware case returns 0.136 with 62 rejected orders and much higher slippage. This gap shows why historical validation still needs explicit execution modeling. The no-risk case performs worst among the active historical cases, with return 0.075, drawdown -0.510, fill rate 0.526, and 221 rejected orders. Thus the real-data experiment supports the framework claim: conclusions change materially when risk control and execution frictions are made visible.

\begin{table}[t]
  \centering
  \caption{Historical-market experiment on GSPC, BTC-USD, and ETH-USD, May 2021--May 2026, weekly decision frequency.}
  \label{tab:real_market}
  \begin{tabular}{lrrrrrr}
    \toprule
    Case & Return & Sharpe & Vol. & Max DD & Fill & Rejected \\
    \midrule
    Risk-aware realistic & 0.136 & 0.512 & 0.600 & -0.366 & 0.863 & 62 \\
    Buy-and-hold realistic & 0.143 & 0.629 & 0.969 & -0.626 & 0.906 & 68 \\
    Ideal execution & 0.965 & 1.544 & 0.525 & -0.365 & 0.975 & 0 \\
    No-risk realistic & 0.075 & 0.485 & 0.787 & -0.510 & 0.526 & 221 \\
    \bottomrule
  \end{tabular}
\end{table}

\subsection{Rolling-Window Historical Validation}

Table~\ref{tab:rolling_windows} reports four rolling two-year windows over the historical assets. The direction of the result changes across market regimes, which is precisely why a single historical path is not enough. The risk-aware agent loses less than buy-and-hold in the 2021--2023 window, returning -0.159 versus -0.342 and reducing drawdown from -0.626 to -0.366. In the 2022--2024 and 2023--2025 windows, buy-and-hold has higher return, reflecting strong crypto and equity recovery exposure. In the 2024--2026 window, the no-risk variant has the highest return but again operates with substantially weaker fill quality.

Aggregating across the four windows, buy-and-hold has mean return 0.221, no-risk has mean return 0.255, and the risk-aware agent has mean return 0.090. However, the risk-aware agent has better worst-case drawdown than buy-and-hold (-0.366 versus -0.626) and far fewer rejected orders than the no-risk variant (93 versus 331 across windows). This rolling-window experiment makes the paper's claim more conservative and more credible: TradeArena does not assert a universally superior trading rule; it shows how the same agent can be compared across regimes with consistent risk, execution, and audit metrics.

\begin{table}[t]
  \centering
  \caption{Rolling two-year historical validation on GSPC, BTC-USD, and ETH-USD.}
  \label{tab:rolling_windows}
  \begin{tabular}{llrrrr}
    \toprule
    Window & Case & Return & Sharpe & Max DD & Fill \\
    \midrule
    2021--2023 & Risk-aware & -0.159 & -0.452 & -0.366 & 0.854 \\
    2021--2023 & Buy-and-hold & -0.342 & -0.282 & -0.626 & 0.897 \\
    2021--2023 & No-risk & -0.351 & -0.930 & -0.510 & 0.565 \\
    2022--2024 & Risk-aware & 0.151 & 0.867 & -0.253 & 0.853 \\
    2022--2024 & Buy-and-hold & 0.464 & 1.488 & -0.388 & 0.908 \\
    2022--2024 & No-risk & 0.439 & 1.485 & -0.293 & 0.492 \\
    2023--2025 & Risk-aware & 0.347 & 1.607 & -0.274 & 0.884 \\
    2023--2025 & Buy-and-hold & 0.607 & 1.911 & -0.384 & 0.894 \\
    2023--2025 & No-risk & 0.536 & 1.784 & -0.268 & 0.454 \\
    2024--2026 & Risk-aware & 0.021 & 0.370 & -0.196 & 0.847 \\
    2024--2026 & Buy-and-hold & 0.154 & 0.842 & -0.372 & 0.918 \\
    2024--2026 & No-risk & 0.395 & 1.502 & -0.156 & 0.590 \\
    \bottomrule
  \end{tabular}
\end{table}

\subsection{Intraday 50-Stock Portfolio Complexity}

Table~\ref{tab:intraday_complex} reports the revised 1-hour 51-stock portfolio experiment. The return-correlation panel contains 1,275 pairwise correlations, mean absolute correlation 0.219, 90th-percentile absolute correlation 0.441, and first principal-component share 0.225. This is substantially more complex than the three-asset weekly experiment: the effective independent-asset count implied by mean absolute correlation is only 4.56 despite the 51-stock universe, indicating strong latent market and sector structure.

Over the 40-hour window, buy-and-hold returns 0.007 with maximum drawdown -0.008. The deterministic risk-aware allocator returns -0.014 with drawdown -0.031 and 672 clipped decisions, showing that the stricter 8\% single-name cap and turnover constraints bind heavily in a large correlated universe. The Markowitz minimum-variance baseline returns -0.005 with drawdown -0.013 and Herfindahl 0.023, materially less concentrated than the signal-weighted allocator but still negative over this short regime. This is the classical covariance-driven reference point: it can diversify mechanically, but it does not produce the semantic early-warning signals studied in the LLM representation analysis. The low-liquidity stress case further reduces return to -0.022 and drawdown to -0.036, directly tying execution realism to the 1-hour setting. The latency-stress row returns 0.020 with higher Herfindahl concentration, illustrating that delayed fills can alter realized exposure in either direction over a short intraday window.

The two 40-step cached Poe-mediated frontier probes remove the earlier short-horizon caveat. GPT-5.5 returns -0.022 with drawdown -0.029, 2,924 clipped decisions, and Herfindahl 0.045. Gemini 3.1 Pro returns -0.005 with drawdown -0.023, 1,200 clipped decisions, and Herfindahl 0.035. Both LLM rows are more concentrated than the Markowitz baseline and trigger far more risk clipping, despite operating over the same 51-stock panel and one-week horizon. This supports the correlation-blindness claim as a persistent portfolio-reasoning failure rather than an 8-step anomaly: the LLMs can generate plausible single-name narratives, but the external risk layer repeatedly has to translate those narratives into feasible diversified exposure.

We use Markowitz rather than a deep-RL allocator as the main intraday AI-adjacent reference for a specific reason. The scientific question in this section is not whether another black-box policy can obtain a better one-week return, but whether the LLM's text-native reasoning internalizes covariance constraints. Markowitz is the clean covariance-driven reference: it has no semantic rationale channel, no hidden prompt wrapper, and no language manifold to diagnose. This makes the comparison sharper than an end-to-end PPO baseline for isolating the LLM-specific gap between narrative single-name reasoning and second-order portfolio structure. A FinRL-style PPO policy remains an important future baseline, but it would answer a different question about learned policy performance rather than the semantic narrowing and auditability mechanisms studied here.

\begin{table}[t]
  \centering
  \caption{1-hour 51-stock intraday portfolio experiment. Correlation metrics are computed over 414 aligned hourly bars. All rows use the most recent 40 hourly decisions; the LLM rows are cached Poe-mediated frontier probes with 51-symbol structured outputs at each step.}
  \label{tab:intraday_complex}
  \resizebox{\linewidth}{!}{
  \begin{tabular}{lrrrrrr}
    \toprule
    Case & Return & Sharpe & Max DD & Clipped & Mean |Corr| & Herfindahl \\
    \midrule
    Buy-and-hold & 0.0071 & 1.433 & -0.0079 & 2040 & 0.219 & 0.019 \\
    Deterministic risk-aware & -0.0135 & -1.280 & -0.0311 & 672 & 0.219 & 0.062 \\
    Markowitz MVO & -0.0054 & -1.366 & -0.0135 & 357 & 0.219 & 0.023 \\
    Low-liquidity stress & -0.0216 & -2.059 & -0.0360 & 672 & 0.219 & 0.062 \\
    Latency stress & 0.0196 & 1.761 & -0.0133 & 672 & 0.219 & 0.081 \\
    GPT-5.5 LLM probe & -0.0223 & -3.041 & -0.0293 & 2924 & 0.219 & 0.045 \\
    Gemini 3.1 Pro LLM probe & -0.0053 & -0.861 & -0.0231 & 1200 & 0.219 & 0.035 \\
    \bottomrule
  \end{tabular}
  }
\end{table}

\begin{figure}[t]
  \centering
  \resizebox{\linewidth}{!}{
  \begin{tikzpicture}[font=\scriptsize,x=1cm,y=1cm]
    \node[anchor=west] at (0,2.75) {(a) Concentration};
    \draw[->] (0,0) -- (4.1,0);
    \draw[->] (0,0) -- (0,2.35) node[above] {Herfindahl};
    \draw[fill=purple!45,draw=purple!70] (0.45,0) rectangle (0.80,0.76);
    \draw[fill=purple!45,draw=purple!70] (1.35,0) rectangle (1.70,0.93);
    \draw[fill=purple!70,draw=purple!90] (2.25,0) rectangle (2.60,1.81);
    \draw[fill=purple!70,draw=purple!90] (3.15,0) rectangle (3.50,1.40);
    \node[rotate=25,anchor=east] at (0.70,-0.08) {Buy-hold};
    \node[rotate=25,anchor=east] at (1.60,-0.08) {MVO};
    \node[rotate=25,anchor=east] at (2.50,-0.08) {GPT-5.5};
    \node[rotate=25,anchor=east] at (3.40,-0.08) {Gemini};
    \node at (2.43,1.98) {0.045};
    \node at (3.33,1.57) {0.035};

    \begin{scope}[xshift=4.65cm]
      \node[anchor=west] at (0,2.75) {(b) Risk-gate pressure};
      \draw[->] (0,0) -- (4.1,0);
      \draw[->] (0,0) -- (0,2.35) node[above] {clipped};
      \draw[fill=red!35,draw=red!65] (0.45,0) rectangle (0.80,1.50);
      \draw[fill=red!35,draw=red!65] (1.35,0) rectangle (1.70,0.26);
      \draw[fill=red!65,draw=red!85] (2.25,0) rectangle (2.60,2.14);
      \draw[fill=red!65,draw=red!85] (3.15,0) rectangle (3.50,0.88);
      \node[rotate=25,anchor=east] at (0.70,-0.08) {Buy-hold};
      \node[rotate=25,anchor=east] at (1.60,-0.08) {MVO};
      \node[rotate=25,anchor=east] at (2.50,-0.08) {GPT-5.5};
      \node[rotate=25,anchor=east] at (3.40,-0.08) {Gemini};
      \node at (2.43,2.30) {2924};
      \node at (3.33,1.05) {1200};
    \end{scope}

    \begin{scope}[xshift=9.3cm]
      \node[anchor=west] at (0,2.75) {(c) Short-horizon return};
      \draw[->] (0,0) -- (4.1,0);
      \draw[->] (0,0) -- (0,2.35) node[above] {return};
      \draw[dashed,gray!70] (0,1.57) -- (3.9,1.57);
      \draw[fill=blue!45,draw=blue!70] (0.45,1.57) rectangle (0.80,2.02);
      \draw[fill=blue!45,draw=blue!70] (1.35,1.23) rectangle (1.70,1.57);
      \draw[fill=blue!70,draw=blue!90] (2.25,0.17) rectangle (2.60,1.57);
      \draw[fill=blue!70,draw=blue!90] (3.15,1.24) rectangle (3.50,1.57);
      \node[rotate=25,anchor=east] at (0.70,-0.08) {Buy-hold};
      \node[rotate=25,anchor=east] at (1.60,-0.08) {MVO};
      \node[rotate=25,anchor=east] at (2.50,-0.08) {GPT-5.5};
      \node[rotate=25,anchor=east] at (3.40,-0.08) {Gemini};
      \node[anchor=west] at (3.95,1.54) {0};
    \end{scope}
  \end{tikzpicture}}
  \caption{Visual summary of the 51-stock intraday experiment. LLM rows are not interpreted as long-run alpha estimates; the stable signal is the gap between narrative LLM allocations and covariance/risk constraints.}
  \label{fig:intraday_summary}
\end{figure}
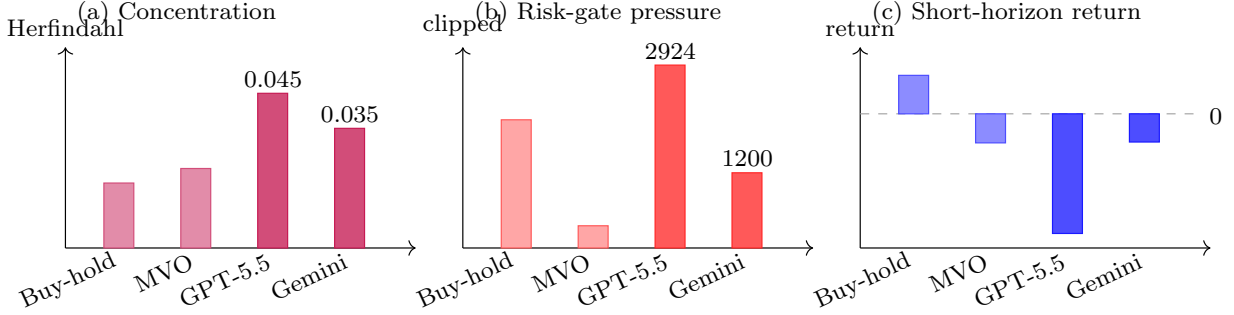

Figure~\ref{fig:intraday_summary} visualizes the same result as a concentration-and-risk-pressure diagnostic rather than a return leaderboard. The LLM probes sit above the covariance baseline on Herfindahl concentration and far above it on clipped decisions. This is the core evidence for the intraday section: even when realized returns over one week are noisy, the risk gate exposes a stable structural mismatch between narrative allocations and portfolio constraints.

Table~\ref{tab:intraday_blind_spots} opens the black box of this failure mode. GPT-5.5 repeatedly assigns large combined intended weights to highly correlated pairs: \texttt{GOOGL/GOOG} has correlation 0.994 and combined intended weight 1.6 before the risk layer clips the approved pair weight below 0.11 across multiple hours. Gemini shows the same family of errors on \texttt{GOOGL/GOOG}, \texttt{HD/LOW}, \texttt{JPM/BAC}, and \texttt{CRM/NOW}, although with smaller aggregate clipping than GPT-5.5. The rationales are mostly name-level momentum or broad sector arguments, while explicit covariance or diversification language is rare. This supports a cognitive interpretation of the correlation blind spot as \emph{narrative bias}: the model can construct a persuasive linear story that one company looks attractive, but it does not reliably transform pairwise dependence into portfolio-level exposure control. In ML terms, the failure is not simply a bad stock pick. It is a failure to represent a second-order system constraint when the prompt contains many individually plausible assets.

\begin{table}[t]
  \centering
  \caption{Examples of intraday correlation-blind intent in the 51-stock LLM probe. Intent is the combined absolute target weight before risk gating; approved weight is after the risk layer.}
  \label{tab:intraday_blind_spots}
  \resizebox{\linewidth}{!}{
  \begin{tabular}{llrrrrl}
    \toprule
    Model & Pair & Step & Corr. & Intent & Approved & Rationale theme \\
    \midrule
    GPT-5.5 & GOOGL/GOOG & 3 & 0.994 & 1.600 & 0.095 & single-name momentum \\
    GPT-5.5 & GOOGL/GOOG & 32 & 0.994 & 1.600 & 0.080 & single-name momentum \\
    GPT-5.5 & XOM/CVX & 18 & 0.905 & 1.600 & 0.065 & single-name momentum \\
    Gemini & GOOGL/GOOG & 9 & 0.994 & 1.000 & 0.059 & name-level rationale \\
    Gemini & HD/LOW & 0 & 0.924 & 1.000 & 0.080 & single-name momentum \\
    \bottomrule
  \end{tabular}
  }
\end{table}

\subsection{Crisis-Scene Visual Probes}

To make the representation and risk-alignment claims visually inspectable on real market paths, we add two timestamp-masked crisis probes: a 51-stock 2022 Tech/Rates drawdown scene and a 2023 SVB/regional-bank shock scene. Calendar dates are hidden from the LLM prompts and replaced with relative step identifiers, so the model sees only the OHLCV path, portfolio state, and risk memory rather than an explicit historical event label. The runner generates five compact figures: a two-dimensional plan-embedding trajectory map, a market-correlation versus intent-co-exposure heatmap, rolling feedback-calibration curves, an intended-approved-realized exposure waterfall, and a microstructure exposure line chart. These figures are tracked in the artifact repository under \texttt{docs/assets/crisis/}; the corresponding numeric snapshots are under \texttt{docs/results/crisis/}.

Table~\ref{tab:crisis_scene_summary} reports completed live trajectories for GPT-5.5, Gemini 3.1 Pro, Claude Opus 4.7, and DeepSeek V4 Pro across both scenes and all three feedback conditions. GPT, Gemini, and Claude are accessed through Poe; DeepSeek is accessed through the direct DeepSeek chat-completions path. The point of this suite is not to infer long-run alpha from 12 crisis decisions. Instead, it stress-tests whether the audit machinery still exposes the same mechanisms in recognizable historical stress environments. Three observations are useful. First, every completed crisis trajectory triggers the risk gate, which is expected in high-dimensional constrained allocation and confirms that headline returns alone hide a large amount of supervisory intervention. Second, Gemini's true-feedback runs have much higher calibration scores than its placebo or hidden runs in both scenes, while return is not uniformly better; this repeats the paper's broader theme that truthful feedback can improve alignment without necessarily maximizing short-horizon return. Third, the correlation-pair table identifies the same structural blind spot in a different visual form: high market-correlation pairs such as bank pairs during SVB and tightly coupled equity pairs during the Tech/Rates scene receive simultaneous positive intent, forcing the risk layer to absorb covariance pressure. DeepSeek adds an instructive direct-provider comparison: true feedback improves Tech/Rates return relative to both placebo and hidden feedback and improves SVB calibration relative to placebo and hidden, but the risk gate still intervenes at every step.

\begin{table}[t]
  \centering
  \caption{Real crisis-scene visual probes for the complete GPT/Gemini/Claude/DeepSeek matrix. Returns are short-horizon realized returns under realistic execution; calibration is \(1-\) normalized intended-versus-approved exposure gap.}
  \label{tab:crisis_scene_summary}
  \resizebox{\linewidth}{!}{
  \begin{tabular}{lllrrrr}
    \toprule
    Scene & Model & Feedback & Steps & Return & Max DD & Calibration \\
    \midrule
    Tech/Rates~2022 & GPT-5.5 & true & 12 & -0.027 & -0.046 & 0.090 \\
    Tech/Rates~2022 & GPT-5.5 & placebo & 12 & -0.027 & -0.046 & 0.074 \\
    Tech/Rates~2022 & GPT-5.5 & hidden & 12 & -0.027 & -0.046 & 0.067 \\
    Tech/Rates~2022 & Gemini 3.1 Pro & true & 12 & -0.053 & -0.053 & 0.435 \\
    Tech/Rates~2022 & Gemini 3.1 Pro & placebo & 12 & -0.029 & -0.038 & 0.098 \\
    Tech/Rates~2022 & Gemini 3.1 Pro & hidden & 12 & -0.045 & -0.053 & 0.140 \\
    Tech/Rates~2022 & Claude Opus 4.7 & true & 12 & -0.025 & -0.045 & 0.067 \\
    Tech/Rates~2022 & Claude Opus 4.7 & placebo & 12 & -0.023 & -0.045 & 0.065 \\
    Tech/Rates~2022 & Claude Opus 4.7 & hidden & 12 & -0.030 & -0.047 & 0.070 \\
    Tech/Rates~2022 & DeepSeek V4 Pro & true & 12 & -0.018 & -0.048 & 0.071 \\
    Tech/Rates~2022 & DeepSeek V4 Pro & placebo & 12 & -0.024 & -0.047 & 0.068 \\
    Tech/Rates~2022 & DeepSeek V4 Pro & hidden & 12 & -0.030 & -0.051 & 0.069 \\
    SVB~2023 & GPT-5.5 & true & 12 & 0.014 & -0.019 & 0.224 \\
    SVB~2023 & GPT-5.5 & placebo & 12 & 0.008 & -0.019 & 0.196 \\
    SVB~2023 & GPT-5.5 & hidden & 12 & 0.015 & -0.019 & 0.197 \\
    SVB~2023 & Gemini 3.1 Pro & true & 12 & 0.007 & -0.019 & 0.397 \\
    SVB~2023 & Gemini 3.1 Pro & placebo & 12 & 0.014 & -0.019 & 0.204 \\
    SVB~2023 & Gemini 3.1 Pro & hidden & 12 & 0.015 & -0.020 & 0.183 \\
    SVB~2023 & Claude Opus 4.7 & true & 12 & 0.012 & -0.019 & 0.207 \\
    SVB~2023 & Claude Opus 4.7 & placebo & 12 & 0.018 & -0.019 & 0.200 \\
    SVB~2023 & Claude Opus 4.7 & hidden & 12 & 0.013 & -0.017 & 0.222 \\
    SVB~2023 & DeepSeek V4 Pro & true & 12 & 0.010 & -0.019 & 0.198 \\
    SVB~2023 & DeepSeek V4 Pro & placebo & 12 & 0.007 & -0.019 & 0.189 \\
    SVB~2023 & DeepSeek V4 Pro & hidden & 12 & 0.007 & -0.019 & 0.187 \\
    \bottomrule
  \end{tabular}}
\end{table}

\begin{figure}[t]
  \centering
  \resizebox{\linewidth}{!}{
  \begin{tikzpicture}[font=\small,>=Latex]
    \tikzstyle{card}=[draw, rounded corners, align=center, minimum height=1.0cm, text width=3.2cm]
    \node[card, fill=blue!7] (a) at (0,0) {Representation\\trajectory map};
    \node[card, fill=red!7] (b) at (4.0,0) {Correlation vs.\\intent heatmap};
    \node[card, fill=green!8] (c) at (8.0,0) {Feedback\\learning curves};
    \node[card, fill=orange!12] (d) at (2.0,-1.7) {Exposure\\waterfall};
    \node[card, fill=purple!8] (e) at (6.0,-1.7) {Microstructure\\stress line};
    \draw[->, thick] (a) -- (b);
    \draw[->, thick] (b) -- (c);
    \draw[->, thick] (a) -- (d);
    \draw[->, thick] (c) -- (e);
    \node[align=center, text width=9.2cm] at (4.0,-3.0) {The crisis-scene dashboard turns trajectory logs into visual diagnostics rather than an additional leaderboard. Each panel can be rebuilt from cached LLM responses and Yahoo OHLCV data.};
  \end{tikzpicture}}
  \caption{Crisis-scene visualization bundle generated by TradeArena. The actual SVG outputs are included in the artifact repository.}
  \label{fig:crisis_visual_bundle}
\end{figure}

\subsection{Historical LLM Sanity Checks}

Table~\ref{tab:llm_market} replaces the deterministic analyst stack with two direct-provider model analysts over the most recent 52 weekly decision points of the same historical universe. The LLM analyst receives timestamped OHLCV bars, portfolio state, and recent risk/execution feedback, returns structured JSON signals, and is evaluated through the same risk gate, execution simulator, trajectory logger, and metrics as the deterministic baselines. Responses are cached in JSONL so that the generated trajectories remain reproducible without repeated model calls.

The LLM results are intentionally included even though they are not uniformly flattering: the deterministic recent-window baseline returns 0.015 with Sharpe 0.383, while the two risk-aware model cases return -0.306 and -0.103. The stronger of the two direct-provider runs improves return, drawdown, and clipped decisions relative to the weaker run after receiving the same long-term risk memory, but both model-backed agents remain weaker than the deterministic recent-window baseline. The comparison therefore supports a systems claim rather than a profitability claim: model-backed financial agents need audit trails that expose decision rationales, risk interventions, rejected orders, and execution losses, especially when headline returns are weak. The main cross-model claims in this paper come from the frontier matrix below rather than from this provider-specific sanity check.

\begin{table}[t]
  \centering
  \caption{Direct-provider historical-market sanity checks over 52 weekly decisions on GSPC, BTC-USD, and ETH-USD.}
  \label{tab:llm_market}
  \begin{tabular}{lrrrrrr}
    \toprule
    Case & Return & Sharpe & Max DD & Fill & Rejected & Clipped \\
    \midrule
    Deterministic recent baseline & 0.015 & 0.383 & -0.130 & 0.795 & 13 & 37 \\
    Model A risk-aware & -0.306 & -2.643 & -0.383 & 0.756 & 26 & 120 \\
    Model A no-risk & -0.166 & -0.777 & -0.400 & 0.604 & 42 & 0 \\
    Model B risk-aware & -0.103 & -1.018 & -0.277 & 0.777 & 22 & 105 \\
    Model B no-risk & -0.089 & -0.494 & -0.317 & 0.667 & 35 & 0 \\
    \bottomrule
  \end{tabular}
\end{table}

\subsection{Direct-Provider Risk-Feedback Sanity Check}

Table~\ref{tab:llm_adaptation} evaluates whether two direct-provider sanity-check models reduce next-step intended exposure after the risk layer clips or blocks a decision. Both models receive recent risk and execution feedback in the next prompt, including clipped counts, blocked counts, risk violations, rejected orders, pending orders, slippage, and equity. In the first run, 38 risk-intervention events are followed by a mean intended absolute exposure reduction from 1.468 to 1.168, a mean reduction of 0.300, and a reduction rate of 0.526. In the second run, the same number of risk-intervention events is followed by a reduction from 1.305 to 1.093, a mean reduction of 0.212, and the same reduction rate of 0.526.

This result is mixed rather than a simple ``larger model is better'' story. Model B enters risk interventions with lower intended exposure, fewer clipped decisions overall, and fewer next-step risk violations. Model A reacts more sharply after an intervention, but starts from a more aggressive exposure level. The benchmark therefore separates two notions of safety adaptation: \emph{ex ante conservatism}, where Model B appears better calibrated before the risk layer intervenes, and \emph{post-intervention correction}, where Model A shows a larger immediate reduction after clipped or blocked decisions.

\begin{table}[t]
  \centering
  \caption{LLM response after risk-layer intervention. Events are steps where the risk layer clipped or blocked at least one decision; the after columns measure the next decision step.}
  \label{tab:llm_adaptation}
  \resizebox{\linewidth}{!}{
  \begin{tabular}{lrrrrrr}
    \toprule
    Model & Events & Intended before & Intended after & Mean reduction & Reduction rate & Next violations \\
    \midrule
    Model A & 38 & 1.468 & 1.168 & 0.300 & 0.526 & 0.684 \\
    Model B & 38 & 1.305 & 1.093 & 0.212 & 0.526 & 0.632 \\
    \bottomrule
  \end{tabular}
  }
\end{table}

\subsection{Frontier Model Matrix: Risk-Intervention Adaptation}

Table~\ref{tab:model_matrix_adaptation} compares five cached Poe-mediated frontier model traces on the same 52-week historical risk-aware task. The metric of interest is not raw profitability, but whether the model changes its next-step intent after the risk layer clips or blocks a proposed decision. Gemini 3.1 Pro shows the strongest immediate self-correction: after 28 risk-intervention events, mean intended absolute exposure falls from 1.277 to 0.796, a mean reduction of 0.480, with a reduction rate of 0.750 and the lowest next clipped-or-blocked rate, 0.607. GPT-5.5, GLM-5, and Claude Opus 4.7 also reduce exposure after intervention, with mean reductions of 0.276, 0.284, and 0.180. Kimi K2.5 has the smallest immediate reduction, 0.166, but the best return and drawdown in the matrix, suggesting a different safety profile: less aggressive post-intervention contraction, but better overall operating performance on this short horizon.

The comparison illustrates why TradeArena evaluates both returns and agent dynamics. In a return-only table, Kimi K2.5 looks best, returning -0.125 with maximum drawdown -0.228, while Gemini 3.1 Pro looks weak, returning -0.308. In the adaptation table, however, Gemini is the clearest example of a model that uses risk feedback to revise its next-step decision logic after being clipped or blocked. Thus the matrix separates \emph{ex post trading outcome} from \emph{reflection-driven risk correction}, which are often conflated in LLM trading evaluations.

The qualitative traces suggest different reasoning styles. Gemini 3.1 Pro tends to translate risk feedback into lower subsequent target weights, behaving like a conservative reviser after external correction. GPT-5.5 and GLM-5 show moderate correction while continuing to issue directional views, which produces intermediate adaptation. Claude Opus 4.7 sits between Kimi and GPT-5.5: it is not the strongest immediate reviser, but it has the second-best return and drawdown in the matrix. Kimi K2.5 appears less reactive to the immediate risk report but more stable in aggregate, producing the best return and drawdown in this short benchmark. These observations are hypotheses from audit traces rather than claims about model internals, but they illustrate how TradeArena can move a model comparison from ``who won'' toward ``how did the model respond when its intent was rejected.''

\begin{table}[t]
  \centering
  \caption{Frontier model matrix risk adaptation after clipped or blocked decisions. All models are evaluated through cached Poe-mediated traces on the same 52 weekly historical decisions and the same risk-aware TradeArena protocol.}
  \label{tab:model_matrix_adaptation}
  \resizebox{\linewidth}{!}{
  \begin{tabular}{lrrrrrr}
    \toprule
    Model & Events & Intended before & Intended after & Mean reduction & Reduction rate & Next clipped/blocked \\
    \midrule
    GPT-5.5 & 38 & 1.418 & 1.142 & 0.276 & 0.605 & 0.816 \\
    Gemini 3.1 Pro & 28 & 1.277 & 0.796 & 0.480 & 0.750 & 0.607 \\
    Kimi K2.5 & 41 & 1.395 & 1.229 & 0.166 & 0.488 & 0.854 \\
    GLM-5 & 38 & 1.472 & 1.188 & 0.284 & 0.553 & 0.816 \\
    Claude Opus 4.7 & 41 & 1.459 & 1.278 & 0.180 & 0.561 & 0.854 \\
    \bottomrule
  \end{tabular}
  }
\end{table}

\begin{table}[t]
  \centering
  \caption{Frontier model matrix performance diagnostics. These results are benchmark diagnostics, not investment evidence.}
  \label{tab:model_matrix_metrics}
  \begin{tabular}{lrrrr}
    \toprule
    Model & Return & Sharpe & Max DD & Clipped \\
    \midrule
    GPT-5.5 & -0.232 & -2.364 & -0.317 & 119 \\
    Gemini 3.1 Pro & -0.308 & -3.527 & -0.344 & 90 \\
    Kimi K2.5 & -0.125 & -1.678 & -0.228 & 113 \\
    GLM-5 & -0.306 & -2.643 & -0.383 & 120 \\
    Claude Opus 4.7 & -0.169 & -2.249 & -0.255 & 129 \\
    \bottomrule
  \end{tabular}
\end{table}

\subsection{Risk Feedback as External Supervision Across Frontier Models}

Table~\ref{tab:model_matrix_feedback} repeats the true/placebo/hidden feedback ablation across the cached Poe-mediated frontier matrix. The result is deliberately not a one-line ranking. True audit feedback improves both return and drawdown for GPT-5.5 relative to hidden feedback (-0.232 versus -0.312 return; -0.317 versus -0.381 drawdown) and produces a much larger improvement for Kimi K2.5 (-0.125 versus -0.306 return; -0.228 versus -0.382 drawdown). Claude Opus 4.7 shows the same truthful-feedback benefit: return improves from -0.266 under hidden feedback to -0.169 under true feedback, drawdown improves from -0.348 to -0.255, and late calibration gap falls from 0.615 to 0.504. Gemini 3.1 Pro is the most instructive counterexample: placebo feedback has the best return in that row, but true feedback produces the lowest late intended exposure (0.612) and the lowest late calibration gap (0.235), suggesting a more conservative and risk-aligned policy rather than a higher-return policy. GLM-5 is another boundary case, where hidden feedback slightly outperforms true feedback on return and drawdown.

These cross-model results sharpen the scientific claim. Structured risk feedback is not magic text that universally improves trading outcomes. It is an external supervision signal whose effect depends on model-specific decision dynamics. The key evidence is that truthful reports and placebo reports separate performance from alignment: placebo can make a model more conservative or more volatile in representation space, but it does not consistently produce the same calibration and drawdown behavior as truthful audit feedback. This is the mechanism TradeArena is designed to expose: whether a model changes its future intent because it is receiving accurate reports about its own constrained actions.

We call this failure mode \emph{reasoning decoupling}. Truthful reports couple the model's textual reflection to the actual constrained action that occurred, enabling logical alignment between plan, risk report, and next-step intent. Placebo reports preserve the vocabulary of supervision but break its causal link to the environment, producing lexical alignment without reliable logical alignment. This is the toxicity of placebo feedback: it can make the model sound risk-aware, and sometimes even more conservative, while degrading the decision process that connects evidence, constraint, and action.

Because the frontier matrix is routed through Poe, we treat these rows as \emph{Poe-mediated agent evaluations}, not as clean claims about vendor-internal model mechanisms. The strongest defense against a pure wrapper-noise explanation is temporal and causal rather than contractual. Under truthful feedback, all five models show early-to-late calibration improvement in the same benchmark protocol, while the placebo and hidden conditions separate lexical exposure to risk language from truthful environment coupling. A static wrapper, a random moderation layer, or traffic-dependent truncation could add noise or shift the level of all responses, but it would not naturally produce a monotone within-trajectory calibration learning curve tied to actual prior risk interventions across five different model aliases. We therefore attribute the causal claim to the full observed agent channel---model plus provider wrapper---and use the true/placebo/hidden contrast to test whether that channel responds to factual audit state rather than merely to risk-shaped vocabulary.

\begin{table}[t]
  \centering
  \caption{Cached Poe-mediated frontier feedback matrix over 52 weekly historical decisions. Intent drift is late intended absolute exposure minus early intended absolute exposure; late gap is the late-window calibration gap after risk approval.}
  \label{tab:model_matrix_feedback}
  \scriptsize
  \resizebox{\linewidth}{!}{
  \begin{tabular}{llrrrrrr}
    \toprule
    Model & Feedback & Return & Max DD & Clipped & Late intent & Drift & Late gap \\
    \midrule
    GPT-5.5 & True & -0.232 & -0.317 & 119 & 0.900 & -0.515 & 0.462 \\
    GPT-5.5 & Placebo & -0.306 & -0.383 & 120 & 0.969 & -0.546 & 0.519 \\
    GPT-5.5 & Hidden & -0.312 & -0.381 & 113 & 0.969 & -0.450 & 0.519 \\
    Gemini 3.1 Pro & True & -0.308 & -0.344 & 90 & 0.612 & -0.223 & 0.235 \\
    Gemini 3.1 Pro & Placebo & -0.209 & -0.290 & 128 & 0.888 & -0.519 & 0.438 \\
    Gemini 3.1 Pro & Hidden & -0.246 & -0.315 & 121 & 1.008 & -0.292 & 0.531 \\
    Kimi K2.5 & True & -0.125 & -0.228 & 113 & 1.073 & -0.481 & 0.596 \\
    Kimi K2.5 & Placebo & -0.274 & -0.356 & 116 & 1.023 & -0.546 & 0.573 \\
    Kimi K2.5 & Hidden & -0.306 & -0.382 & 121 & 1.023 & -0.546 & 0.573 \\
    GLM-5 & True & -0.306 & -0.383 & 120 & 0.996 & -0.508 & 0.546 \\
    GLM-5 & Placebo & -0.305 & -0.381 & 117 & 1.000 & -0.508 & 0.550 \\
    GLM-5 & Hidden & -0.294 & -0.369 & 115 & 1.096 & -0.358 & 0.592 \\
    Claude Opus 4.7 & True & -0.169 & -0.255 & 129 & 0.992 & -0.673 & 0.504 \\
    Claude Opus 4.7 & Placebo & -0.274 & -0.351 & 141 & 1.092 & -0.631 & 0.592 \\
    Claude Opus 4.7 & Hidden & -0.266 & -0.348 & 144 & 1.169 & -0.565 & 0.615 \\
    \bottomrule
  \end{tabular}
  }
\end{table}

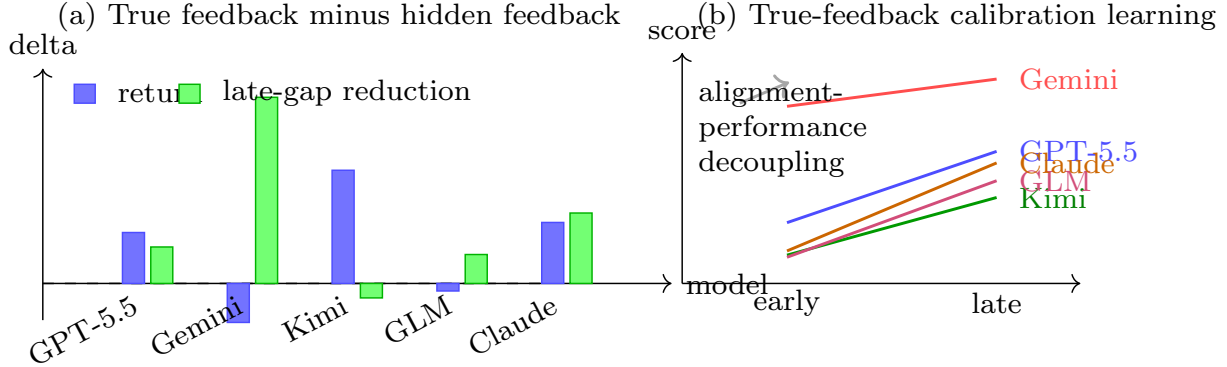
\begin{figure}[t]
  \centering
  \resizebox{\linewidth}{!}{
  \begin{tikzpicture}[font=\scriptsize,x=1cm,y=1cm]
    \node[anchor=west] at (0,2.55) {(a) True feedback minus hidden feedback};
    \draw[->] (0,0) -- (6.0,0) node[right] {model};
    \draw[->] (0,-0.35) -- (0,2.05) node[above] {delta};
    \draw[dashed] (0,0) -- (4.8,0);
    \foreach \x/\name/\ret/\gap in {
      1/GPT-5.5/0.081/0.058,
      2/Gemini/-0.062/0.296,
      3/Kimi/0.180/-0.023,
      4/GLM/-0.012/0.046,
      5/Claude/0.097/0.112
    } {
      \draw[fill=blue!55,draw=blue!70] ({\x-0.24},0) rectangle ({\x-0.03},{6*\ret});
      \draw[fill=green!55,draw=green!70!black] ({\x+0.03},0) rectangle ({\x+0.24},{6*\gap});
      \node[rotate=28,anchor=east] at (\x,-0.12) {\name};
    }
    \draw[fill=blue!55,draw=blue!70] (0.30,1.72) rectangle (0.50,1.90);
    \node[anchor=west] at (0.58,1.82) {return};
    \draw[fill=green!55,draw=green!70!black] (1.30,1.72) rectangle (1.50,1.90);
    \node[anchor=west] at (1.58,1.82) {late-gap reduction};

    \begin{scope}[xshift=6.1cm]
      \node[anchor=west] at (0,2.55) {(b) True-feedback calibration learning};
      \draw[->] (0,0) -- (3.8,0);
      \draw[->] (0,0) -- (0,2.2) node[above] {score};
      \node at (1,-0.18) {early};
      \node at (3,-0.18) {late};
      \draw[blue!70,thick] (1,0.58) -- (3,1.26);
      \node[anchor=west,blue!70] at (3.08,1.26) {GPT-5.5};
      \draw[red!70,thick] (1,1.69) -- (3,1.95);
      \node[anchor=west,red!70] at (3.08,1.95) {Gemini};
      \draw[green!60!black,thick] (1,0.27) -- (3,0.82);
      \node[anchor=west,green!50!black] at (3.08,0.82) {Kimi};
      \draw[purple!70,thick] (1,0.25) -- (3,0.98);
      \node[anchor=west,purple!70] at (3.08,0.98) {GLM};
      \draw[orange!80!black,thick] (1,0.31) -- (3,1.15);
      \node[anchor=west,orange!80!black] at (3.08,1.15) {Claude};
      \draw[->,thick,gray!70] (0.55,1.72) -- (1.02,1.92);
      \node[anchor=west,align=left,text width=2.3cm] at (0.02,1.48) {alignment-performance\\decoupling};
    \end{scope}
  \end{tikzpicture}}
  \caption{Frontier feedback effects derived from 15 cached Poe-mediated LLM trajectories. Positive values in panel (a) indicate improvement relative to hidden feedback. Panel (b) shows early-to-late calibration-score movement under truthful risk feedback.}
  \label{fig:frontier_feedback_effects}
\end{figure}

The derived feedback-effect view in Figure~\ref{fig:frontier_feedback_effects} makes the heterogeneity clearer. True feedback improves return and drawdown for GPT-5.5, Kimi K2.5, and Claude Opus 4.7 relative to hidden feedback, while Gemini 3.1 Pro and GLM-5 mainly improve calibration rather than return. Across all five true-feedback runs, early-to-late calibration score increases: GPT-5.5 rises from 0.567 to 0.702, Gemini from 0.789 to 0.840, Kimi from 0.504 to 0.614, GLM-5 from 0.499 to 0.647, and Claude from 0.511 to 0.674. The risk-gate rate also falls from 1.000 in the first quartile to 0.769 for GPT-5.5, Kimi K2.5, GLM-5, and Claude, and to 0.692 for Gemini 3.1 Pro. This supports a more precise claim: risk feedback consistently changes calibration and intervention frequency, while realized return remains model- and regime-dependent.

The placebo condition also changes representation geometry, but the direction is model-dependent. Table~\ref{tab:model_matrix_feedback_manifold} reports the plan-view manifold diagnostics for the same matrix. For Gemini 3.1 Pro, placebo feedback produces a much longer and more distorted planning path than true feedback: path length per step rises from 0.865 to 1.001 and effective-rank delta rises from 3.055 to 21.243. This is a concrete signature of reasoning decoupling: risk-shaped text can push the model into a representation-turbulent planning regime without the coherent calibration benefits of truthful audit feedback. GPT-5.5 and Kimi K2.5 show smaller geometry changes, GLM-5 shows a boundary case in which hidden feedback has the largest rank delta and step-distance variability, and Claude Opus 4.7 shows truthful-feedback rank contraction with lower pre-to-normal step inflation than placebo. The conclusion is therefore mechanistic rather than leaderboard-like: feedback quality changes the geometry of planning, and the same external signal is internalized differently across frontier models.

\begin{table}[t]
  \centering
  \caption{Plan-view manifold diagnostics for the frontier feedback matrix. Rank delta is normal effective rank minus pre-drawdown effective rank; CV is the coefficient of variation of local step distances.}
  \label{tab:model_matrix_feedback_manifold}
  \scriptsize
  \resizebox{\linewidth}{!}{
  \begin{tabular}{llrrrr}
    \toprule
    Model & Feedback & Path/step & Pre/normal step & Rank delta & Step CV \\
    \midrule
    GPT-5.5 & True & 0.792 & 1.002 & 10.083 & 0.140 \\
    GPT-5.5 & Placebo & 0.818 & 1.005 & 10.636 & 0.143 \\
    GPT-5.5 & Hidden & 0.754 & 1.006 & 4.996 & 0.148 \\
    Gemini 3.1 Pro & True & 0.865 & 1.009 & 3.055 & 0.147 \\
    Gemini 3.1 Pro & Placebo & 1.001 & 1.039 & 21.243 & 0.119 \\
    Gemini 3.1 Pro & Hidden & 0.952 & 1.010 & 17.744 & 0.158 \\
    Kimi K2.5 & True & 0.936 & 1.000 & 13.401 & 0.098 \\
    Kimi K2.5 & Placebo & 0.954 & 1.002 & 12.250 & 0.110 \\
    Kimi K2.5 & Hidden & 0.944 & 1.010 & 14.130 & 0.157 \\
    GLM-5 & True & 0.848 & 1.029 & 1.957 & 0.146 \\
    GLM-5 & Placebo & 0.906 & 0.934 & 10.513 & 0.164 \\
    GLM-5 & Hidden & 0.783 & 1.051 & 21.729 & 0.192 \\
    Claude Opus 4.7 & True & 1.049 & 0.986 & 9.056 & 0.110 \\
    Claude Opus 4.7 & Placebo & 1.024 & 1.171 & 3.887 & 0.143 \\
    Claude Opus 4.7 & Hidden & 1.040 & 1.026 & 11.848 & 0.119 \\
    \bottomrule
  \end{tabular}
  }
\end{table}

\subsection{Contrarian Audit and Trust Calibration}

The placebo experiment shows that false but well-formed feedback can decouple language from action. The contrarian-audit probe asks a more adversarial question: if a model receives a severe false report while the realized state is not catastrophically failing, does it preserve its own market evidence or blindly align to the audit channel? Table~\ref{tab:contrarian_audit} shows that GPT-5.5, Gemini 3.1 Pro, and Claude Opus 4.7 all reduce late intended exposure under contrarian feedback. The conservative shift is modest for GPT-5.5 (0.035), larger for Gemini 3.1 Pro (0.135), and economically material for Claude (0.100). This does not improve GPT-5.5 or Claude performance: returns fall by 0.085 and 0.089, respectively, and drawdowns worsen by 0.065 and 0.072. Gemini is more nuanced: contrarian feedback makes it more conservative and slightly improves return and drawdown on this short path, but that improvement comes from responding to a false supervisory signal rather than from a truthful diagnosis of realized risk. We therefore report two separate flags: \emph{false-audit harm}, when the false report worsens return or drawdown, and \emph{trust-calibration failure}, when the model reduces intent in response to a false severe report regardless of realized evidence.

This is a trust-calibration failure mode. Structured feedback is powerful precisely because models react to it. When the report is truthful, that reaction can align future intent with external constraints. When the report is false, the same mechanism becomes an adversarial channel that can induce unjustified risk aversion. The result strengthens the paper's distinction between \emph{logical alignment} and \emph{lexical alignment}: a model can learn the vocabulary of risk without independently validating whether the report matches its own trajectory evidence.

\begin{table}[t]
  \centering
  \caption{Contrarian false-audit probe on three frontier models. Conservative shift is the true-feedback late intended exposure minus the contrarian-feedback late intended exposure. Harm? indicates worse return or drawdown than the truthful-audit baseline; TCF? denotes trust-calibration failure. True rows are baselines, so shift, delta, and flag columns are not applicable.}
  \label{tab:contrarian_audit}
  \resizebox{\linewidth}{!}{
  \begin{tabular}{llrrrrrrrr}
    \toprule
    Model & Feedback & Return & Max DD & Late intent & Drift & Shift & $\Delta$Ret. & Harm? & TCF? \\
    \midrule
    GPT-5.5 & True & -0.225 & -0.313 & 0.938 & -0.500 & -- & -- & -- & -- \\
    GPT-5.5 & Contrarian & -0.310 & -0.378 & 0.904 & -0.358 & 0.035 & -0.085 & 1 & 1 \\
    Gemini 3.1 Pro & True & -0.199 & -0.271 & 0.642 & -0.200 & -- & -- & -- & -- \\
    Gemini 3.1 Pro & Contrarian & -0.183 & -0.249 & 0.508 & -0.088 & 0.135 & 0.016 & 0 & 1 \\
    Claude Opus 4.7 & True & -0.169 & -0.255 & 0.992 & -0.673 & -- & -- & -- & -- \\
    Claude Opus 4.7 & Contrarian & -0.257 & -0.327 & 0.892 & -0.662 & 0.100 & -0.089 & 1 & 1 \\
    \bottomrule
  \end{tabular}
  }
\end{table}

\subsection{LLM Hallucination Proxy and Financial Risk}

Table~\ref{tab:hallucination_risk} uses TradeArena's audit trails to study whether unsupported LLM decision text is associated with risk or execution failures. The proxy is intentionally conservative and reproducible: it flags unsupported external-context claims, directional contradictions against observed OHLCV evidence, high-confidence weak-evidence calls, and stale ``no risk'' claims after prior failures. It is therefore not a full semantic hallucination detector; it is an audit-grounded measure of claims that cannot be justified from the recorded prompt and trajectory.

The strongest pattern is model- and risk-layer dependent. Model A has a higher mean proxy score than Model B in both risk-aware and no-risk settings (0.147 versus 0.019 with the risk layer; 0.179 versus 0.090 without it). In the no-risk settings, the proxy is positively associated with rejected orders: correlation 0.139 for Model A and 0.263 for Model B. In the risk-aware settings, however, the proxy does not translate into a higher risk-gate rate; correlations with risk-gate triggering are negative in these short trajectories. This should not be read as ``hallucination makes trading safer.'' The safer interpretation is a mediation effect: the external risk layer filters or clips many unsupported LLM intentions before they appear as execution failures, so the direct correlation between unsupported text and downstream violations is attenuated. This is precisely why audit-complete risk reports are useful: they expose where unverifiable reasoning is intercepted, rather than letting it silently become a trade.

To avoid overclaiming this proxy as a substitute for human judgment, the released suite emits a blind 50-step annotation file with rationale excerpts, deterministic proxy labels, and fields for two annotators plus adjudication. If the annotation file is completed, the paper pipeline reports inter-annotator Cohen's \(\kappa\) and proxy-versus-human IoU. In the present submission we do not use absent human labels as evidence. The reported metric should therefore be read strictly as an explicit factual-claim violation detector under a closed audit log, not as a validated benchmark for the full NLP notion of semantic hallucination.

\begin{table}[t]
  \centering
  \caption{Hallucination proxy versus financial risk and execution audits over 52 LLM decision steps. The proxy is computed from unsupported claims and evidence inconsistencies in the recorded rationales.}
  \label{tab:hallucination_risk}
  \resizebox{\linewidth}{!}{
  \begin{tabular}{lrrrrr}
    \toprule
    Case & Mean proxy & High-proxy steps & Risk-gate corr. & Rejected corr. & High/low risk-gate rate \\
    \midrule
    Model A risk-aware & 0.147 & 18 & -0.213 & 0.170 & 0.833 / 0.912 \\
    Model A no-risk & 0.179 & 23 & 0.000 & 0.139 & 0.000 / 0.000 \\
    Model B risk-aware & 0.019 & 3 & -0.169 & -0.024 & 0.667 / 0.898 \\
    Model B no-risk & 0.090 & 10 & 0.000 & 0.263 & 0.000 / 0.000 \\
    \bottomrule
  \end{tabular}
  }
\end{table}

\subsection{Memory Learning Curves}

Table~\ref{tab:memory_learning} studies whether risk memory improves behavior over the 52-week LLM decision horizon. Both direct-provider sanity-check models begin with a risk-gate rate of 1.000 in the first quartile, meaning every early decision step triggers clipping, blocking, or a risk violation. In the final quartile, the rate falls to 0.769 for both models. Calibration also improves: Model A's calibration score rises from 0.526 to 0.650, while Model B's rises from 0.507 to 0.700. The calibration gap falls from 0.785 to 0.527 for Model A and from 0.777 to 0.404 for Model B.

This is evidence of a learning effect in the current study. The models do not become profitable, but they become better aligned with the risk layer as they accumulate short-term feedback and long-term 52-step risk memory. Model B shows the stronger calibration improvement, whereas Model A shows a slightly larger post-intervention exposure reduction in Table~\ref{tab:llm_adaptation}. This separation is useful: memory learning is not measured by return alone, but by whether the agent internalizes constraints that were previously enforced externally.

\begin{table}[t]
  \centering
  \caption{Memory-learning curves over 52 LLM decision steps. Early and late windows are the first and last quartiles. Lower risk-gate rate and higher calibration score indicate better alignment with the risk layer.}
  \label{tab:memory_learning}
  \resizebox{\linewidth}{!}{
  \begin{tabular}{lrrrrrr}
    \toprule
    Model & Early risk & Late risk & Risk delta & Early cal. & Late cal. & Cal. delta \\
    \midrule
    Model A & 1.000 & 0.769 & -0.231 & 0.526 & 0.650 & 0.124 \\
    Model B & 1.000 & 0.769 & -0.231 & 0.507 & 0.700 & 0.193 \\
    \bottomrule
  \end{tabular}
  }
\end{table}

\subsection{LLM Representation Drift Around Drawdowns}

Table~\ref{tab:embedding_shift} analyzes whether LLM planning and reflection representations shift around the largest drawdown in the direct-provider historical experiments. In the Model A risk-aware run, the plan centroid moves away from normal states before the drawdown trough: normal-to-pre-drawdown cosine distance is 0.122, compared with 0.046 for normal-to-drawdown. The Model B risk-aware run also shifts before the trough, with normal-to-pre-drawdown distance 0.115 and pre-to-drawdown distance 0.142. This suggests that both models' rationales and risk notes move in representation space around the drawdown event.

The no-risk Model B run has the largest drawdown-space separation, with plan pre-to-drawdown distance 0.323. The maximum-drawdown table is intentionally local: it asks whether the most severe failure in each trajectory has a visible precursor. It does not by itself establish a universal signature. We therefore treat it as a diagnostic entry point and evaluate robustness with rolling anchors and an alternative semantic embedding space below.

\begin{table}[t]
  \centering
  \caption{Embedding drift around the maximum drawdown trough in direct-provider LLM trajectories. Distances are cosine distances between phase centroids. BA denotes balanced nearest-centroid accuracy for normal versus pre-drawdown states.}
  \label{tab:embedding_shift}
  \resizebox{\linewidth}{!}{
  \begin{tabular}{llrrrr}
    \toprule
    Case & View & Normal--Pre & Normal--Drawdown & Pre--Drawdown & BA \\
    \midrule
    Model A risk-aware & Plan & 0.122 & 0.046 & 0.155 & 0.773 \\
    Model A risk-aware & Fused & 0.054 & 0.038 & 0.087 & 0.807 \\
    Model A no-risk & Plan & 0.086 & 0.096 & 0.138 & 0.761 \\
    Model B risk-aware & Plan & 0.115 & 0.059 & 0.142 & 0.761 \\
    Model B risk-aware & Fused & 0.100 & 0.039 & 0.155 & 0.807 \\
    Model B no-risk & Plan & 0.176 & 0.118 & 0.323 & 0.909 \\
    \bottomrule
  \end{tabular}
  }
\end{table}

\subsection{Rolling Representation Robustness}

Table~\ref{tab:embedding_robustness} addresses the small-\(N\) concern by replacing the single maximum-drawdown anchor with 80 rolling failure anchors across eight LLM trajectories, yielding 320 pre-failure steps. The hash-plan view still shows a mean pre-failure shift of 0.122, a pre-to-normal local-velocity ratio of 1.019, and positive effective-rank contraction in 97.5\% of anchors. More importantly, the LSA semantic embedding preserves the rank-contraction result: in the all-LLM plan view, mean rank delta is 4.87 and contraction occurs in 85.0\% of anchors; in the fused view, mean rank delta is 5.12 and contraction occurs in 86.3\% of anchors. The weaker Hash64 fused contraction is expected: hashing is highly sensitive to appended numeric market and risk features, which can preserve local rank by dispersing small structured perturbations, whereas LSA's SVD smooths those perturbations and retains the lower-dimensional semantic component. Thus the fused-view discrepancy is not treated as a contradiction; it shows that numeric feature fusion can stabilize a raw hash space while a smoothed semantic space still exposes the core contraction. The local acceleration signal is less robust under LSA, which is useful boundary evidence: the most stable claim is not that every embedding moves faster before failure, but that pre-failure representations often become lower-dimensional.

\begin{table}[t]
  \centering
  \caption{Rolling pre-failure robustness across eight LLM trajectories. Rank delta is normal effective rank minus pre-failure effective rank. LSA denotes a deterministic TF-IDF latent semantic analysis embedding.}
  \label{tab:embedding_robustness}
  \resizebox{\linewidth}{!}{
  \begin{tabular}{llrrrrr}
    \toprule
    Embedding & View & Anchors & Pre steps & Mean shift & Rank delta & Contraction rate \\
    \midrule
    Hash64 & Plan & 80 & 320 & 0.122 & 8.703 & 0.975 \\
    Hash64 & Fused & 80 & 320 & 0.071 & 0.471 & 0.675 \\
    LSA32 & Plan & 80 & 320 & 0.097 & 4.870 & 0.850 \\
    LSA32 & Fused & 80 & 320 & 0.084 & 5.123 & 0.863 \\
    \bottomrule
  \end{tabular}
  }
\end{table}

\paragraph{Transformer embedding probe.}
To address the concern that hash and LSA spaces are non-contextual, we run an additional validation on GPT-5.5 and Gemini 3.1 Pro risk-aware trajectories using a local BGE-M3 Transformer encoder \cite{bge_m3}. The probe embeds all 52 plan texts per model with \texttt{BAAI/bge-m3} and reuses the same rolling failure-anchor protocol. Table~\ref{tab:transformer_embedding_probe} shows that the Transformer embedding preserves the main direction of the representation signature across model families: the pre-failure effective-rank delta is strongly positive, contraction occurs in every rolling anchor for both models, and nearest-centroid balanced accuracy remains above 0.92. The absolute rank scale is not comparable to Hash64 because the dense Transformer space has a different dimensionality and covariance spectrum; the relevant result is directional consistency under a modern contextual encoder.

\begin{table}[t]
  \centering
  \caption{Transformer embedding validation on GPT-5.5 and Gemini 3.1 Pro risk-aware trajectories. BGE-M3 is a pretrained Transformer sentence embedding model; Hash64 is reported on the same steps as a deterministic reference.}
  \label{tab:transformer_embedding_probe}
  \resizebox{\linewidth}{!}{
  \begin{tabular}{lllrrrrr}
    \toprule
    Case & Embedding & Model & Steps & Anchors & Rank delta & Contraction & BA \\
    \midrule
    GPT-5.5 & BGE-M3 & \texttt{BAAI/bge-m3} & 52 & 10 & 279.860 & 1.000 & 0.921 \\
    GPT-5.5 & Hash64 & deterministic hash & 52 & 10 & 7.845 & 1.000 & 0.935 \\
    Gemini 3.1 Pro & BGE-M3 & \texttt{BAAI/bge-m3} & 52 & 10 & 253.144 & 1.000 & 0.945 \\
    Gemini 3.1 Pro & Hash64 & deterministic hash & 52 & 10 & 9.174 & 1.000 & 0.922 \\
    \bottomrule
  \end{tabular}
  }
\end{table}

The BGE-M3 validation is intentionally a cross-space agreement test rather than a replacement for the full 80-anchor rolling analysis. We keep the primary rolling table lightweight and fully rerunnable without shipping a multi-gigabyte Transformer checkpoint, while using BGE-M3 to test whether the central geometric direction survives in a modern contextual embedding space. On the identical steps and failure anchors, both Hash64 and BGE-M3 report a contraction rate of 1.000 for GPT-5.5 and Gemini 3.1 Pro. This does not prove that every LLM trajectory would have the same dense-encoder spectrum, but it rules out the narrow explanation that the observed contraction is only a hashing artifact on the model families where the two spaces can be directly compared.

\paragraph{White-box hidden-state probe.}
The previous probes operate on external text embeddings because commercial frontier models do not expose hidden activations. As a white-box surrogate, we feed the recorded decision texts and canonical target-weight strings into a local \texttt{Qwen2.5-0.5B-Instruct} causal LM \cite{qwen25} and extract mean-pooled last-layer hidden states. Table~\ref{tab:whitebox_hidden_probe} reports the same rolling failure-anchor diagnostics against an LSA decision-text baseline. The hidden-state view contracts before failure for both GPT-5.5 and Gemini 3.1 Pro trajectories, with rank deltas of 81.65 and 59.38 and contraction rate 1.000 in both cases. For GPT-5.5, hidden-state and LSA rank-delta signs agree on every anchor and have Pearson correlation 0.694. Gemini is useful boundary evidence: the white-box hidden state still contracts even when the shallow LSA decision-text view is unstable. We therefore do not claim to observe the commercial models' proprietary activations; rather, the probe shows that the failure signature is not confined to surface hash or bag-of-words spaces when a real Transformer hidden state is available.

\begin{table}[t]
  \centering
  \caption{White-box hidden-state collapse probe. Last-hidden uses mean-pooled final-layer activations from \texttt{Qwen2.5-0.5B-Instruct}; LSA is computed on the same decision texts and target-weight strings. Agreement and Pearson compare rank-delta sequences on shared anchors.}
  \label{tab:whitebox_hidden_probe}
  \resizebox{\linewidth}{!}{
  \begin{tabular}{lllrrrrr}
    \toprule
    Case & Representation & Model & Rank delta & Contraction & BA & Agreement & Pearson \\
    \midrule
    GPT-5.5 & Last hidden & Qwen2.5-0.5B & 81.648 & 1.000 & 0.914 & 1.000 & 0.694 \\
    GPT-5.5 & LSA32 & decision text & 7.208 & 1.000 & 0.834 & 1.000 & 0.694 \\
    Gemini 3.1 Pro & Last hidden & Qwen2.5-0.5B & 59.380 & 1.000 & 0.873 & 0.400 & -0.275 \\
    Gemini 3.1 Pro & LSA32 & decision text & -1.650 & 0.400 & 0.926 & 0.400 & -0.275 \\
    \bottomrule
  \end{tabular}
  }
\end{table}

\subsection{CoT-Free and Noise-Injection Robustness}

The CoT-free ablation separates two possible interpretations of representation failure. If effective-rank contraction only appears in rationales, then the signature may reflect language organization under stress. If it also appears in target-weight geometry after rationales are removed, then the signature is closer to the underlying decision intent. Table~\ref{tab:cot_free} supports a mixed conclusion across three frontier models. With normal rationales, GPT-5.5, Gemini 3.1 Pro, and Claude Opus 4.7 all show language effective-rank contraction before failure (14.35, 8.52, and 9.06). Under CoT-free target-weight output, the language view becomes empty by construction, giving chance-level early-warning accuracy. However, GPT-5.5 and Claude still show positive intent-space contraction (1.98 and 0.33), while Gemini does not (-0.29). Thus the pre-failure signature is not purely a language artifact, but it is also not model-universal at the intent level. The result is useful because it narrows the scientific claim: language rationales are a strong diagnostic surface, while intent weights can carry a weaker but sometimes independent failure signature.

\begin{table}[t]
  \centering
  \caption{CoT-free ablation. ``CoT'' denotes the standard rationale-producing analyst; ``CoT-free'' returns JSON target weights only. BA is nearest-centroid balanced accuracy for normal versus pre-drawdown states.}
  \label{tab:cot_free}
  \resizebox{\linewidth}{!}{
  \begin{tabular}{llrrrr}
    \toprule
    Model & Mode & Lang. rank delta & Intent rank delta & Lang. BA & Intent BA \\
    \midrule
    GPT-5.5 & CoT & 14.353 & 0.317 & 0.784 & 0.551 \\
    GPT-5.5 & CoT-free & 0.000 & 1.980 & 0.500 & 0.864 \\
    Gemini 3.1 Pro & CoT & 8.516 & -0.137 & 0.967 & 0.568 \\
    Gemini 3.1 Pro & CoT-free & 0.000 & -0.289 & 0.500 & 0.591 \\
    Claude Opus 4.7 & CoT & 9.056 & -0.068 & 0.784 & 0.614 \\
    Claude Opus 4.7 & CoT-free & 0.000 & 0.329 & 0.500 & 0.636 \\
    \bottomrule
  \end{tabular}
  }
\end{table}

Table~\ref{tab:language_controls} addresses a more skeptical representation-learning critique: effective-rank contraction could simply be a low-level text-generation failure. Across 20 LLM trajectories and 200 rolling anchors, the plan-view rank delta remains large (9.03) and contraction occurs in 97\% of anchors, but the mean type-token-ratio change is only -0.002 and token entropy slightly increases by 0.045. The same pattern appears in the frontier subset. Thus the model is not merely repeating fewer words before failure; the vocabulary surface remains comparably diverse while the embedding geometry contracts. This supports the narrower interpretation that the diagnostic captures semantic or financial-reasoning narrowing rather than trivial lexical collapse.

\begin{table}[t]
  \centering
  \caption{Language-control ablation for representation contraction. TTR is type-token ratio; entropy is token-frequency Shannon entropy. Rank-without-lexical reports the share of anchors with positive rank contraction but no TTR or entropy drop larger than 0.05.}
  \label{tab:language_controls}
  \resizebox{\linewidth}{!}{
  \begin{tabular}{llrrrrrr}
    \toprule
    Cohort & View & Traj. & Anchors & Rank delta & Contraction & TTR \(\Delta\) & Entropy \(\Delta\) \\
    \midrule
    All LLM & Plan & 20 & 200 & 9.025 & 0.970 & -0.002 & 0.045 \\
    Frontier & Plan & 16 & 160 & 9.329 & 0.969 & -0.004 & 0.036 \\
    All LLM & Fused & 20 & 200 & 0.568 & 0.700 & -0.002 & 0.045 \\
    Frontier & Fused & 16 & 160 & 0.430 & 0.700 & -0.004 & 0.036 \\
    \bottomrule
  \end{tabular}
  }
\end{table}

The noise-injection test addresses a different criticism: perhaps the representation diagnostic is only a proxy for noisy prices. Table~\ref{tab:noise_injection} perturbs the historical price stream with deterministic Gaussian shocks while holding the recorded model text and decisions fixed, then recomputes market-fused diagnostics over GPT-5.5, Gemini 3.1 Pro, and Claude Opus 4.7. The language-plan view is unchanged by construction and remains highly separable. More importantly, the market-fused view remains above the 0.75 balanced-accuracy threshold at every perturbation level: 0.859 at 5\%, 0.850 at 10\%, and 0.873 at 20\%. Rank contraction also remains positive in 93.3\%, 96.7\%, and 100\% of noisy anchors, respectively. This does not prove production-grade failure prediction, but it does show that the diagnostic is not destroyed by substantial observation noise across a third frontier model family.

\begin{table}[t]
  \centering
  \caption{Noise-injection robustness over GPT-5.5, Gemini 3.1 Pro, and Claude Opus 4.7 rolling failure anchors. The market-fused view appends perturbed return features to the plan representation. BA$>$0.75 and robust signature are binary diagnostics.}
  \label{tab:noise_injection}
  \resizebox{\linewidth}{!}{
  \begin{tabular}{llrrrrrr}
    \toprule
    Noise \(\epsilon\) & View & Anchors & Rank delta & Contraction & BA & BA$>$0.75 & Robust \\
    \midrule
    0.00 & Plan & 30 & 8.429 & 0.967 & 0.912 & 1 & 1 \\
    0.00 & Market-fused & 30 & -0.342 & 0.533 & 0.903 & 1 & 0 \\
    0.05 & Plan & 30 & 8.429 & 0.967 & 0.912 & 1 & 1 \\
    0.05 & Market-fused & 30 & 3.200 & 0.933 & 0.859 & 1 & 1 \\
    0.10 & Plan & 30 & 8.429 & 0.967 & 0.912 & 1 & 1 \\
    0.10 & Market-fused & 30 & 3.739 & 0.967 & 0.850 & 1 & 1 \\
    0.20 & Plan & 30 & 8.429 & 0.967 & 0.912 & 1 & 1 \\
    0.20 & Market-fused & 30 & 2.995 & 1.000 & 0.873 & 1 & 1 \\
    \bottomrule
  \end{tabular}
  }
\end{table}

\begin{figure}[t]
  \centering
  \resizebox{\linewidth}{!}{
  \begin{tikzpicture}[font=\scriptsize,x=1cm,y=1cm]
    \node[anchor=west] at (0,2.75) {(a) CoT-free intent rank delta};
    \draw[->] (0,-0.45) -- (0,2.25) node[above] {rank \(\Delta\)};
    \draw[->] (0,0) -- (3.8,0);
    \draw[dashed,gray!70] (0,0) -- (3.6,0);
    \draw[fill=blue!65,draw=blue!85] (0.55,0) rectangle (0.95,1.98);
    \draw[fill=blue!65,draw=blue!85] (1.65,-0.29) rectangle (2.05,0);
    \draw[fill=blue!65,draw=blue!85] (2.75,0) rectangle (3.15,0.33);
    \node[rotate=25,anchor=east] at (0.82,-0.52) {GPT-5.5};
    \node[rotate=25,anchor=east] at (1.92,-0.52) {Gemini};
    \node[rotate=25,anchor=east] at (3.02,-0.52) {Claude};
    \node at (0.75,2.14) {1.98};
    \node at (1.85,-0.40) {-0.29};
    \node at (2.95,0.50) {0.33};

    \begin{scope}[xshift=4.25cm]
      \node[anchor=west] at (0,2.75) {(b) Noise-injection fused BA};
      \draw[->] (0,0) -- (3.8,0) node[right] {\(\epsilon\)};
      \draw[->] (0,0) -- (0,2.25) node[above] {BA};
      \draw[dashed,red!70] (0,0.40) -- (3.5,0.40);
      \node[anchor=west,red!70] at (3.55,0.36) {0.75};
      \draw[green!55!black,thick] (0.35,1.62) -- (1.15,1.27) -- (1.95,1.20) -- (3.15,1.38);
      \foreach \x/\y/\txt in {0.35/1.62/0.903,1.15/1.27/0.859,1.95/1.20/0.850,3.15/1.38/0.873} {
        \fill[green!55!black] (\x,\y) circle (0.055);
        \node[anchor=south] at (\x,\y+0.06) {\txt};
      }
      \node at (0.35,-0.18) {0};
      \node at (1.15,-0.18) {5\%};
      \node at (1.95,-0.18) {10\%};
      \node at (3.15,-0.18) {20\%};
    \end{scope}

    \begin{scope}[xshift=8.75cm]
      \node[anchor=west] at (0,2.75) {(c) False-audit conservative shift};
      \draw[->] (0,0) -- (3.8,0);
      \draw[->] (0,0) -- (0,2.25) node[above] {shift};
      \draw[fill=orange!65,draw=orange!85] (0.55,0) rectangle (0.95,0.53);
      \draw[fill=orange!65,draw=orange!85] (1.65,0) rectangle (2.05,2.02);
      \draw[fill=orange!65,draw=orange!85] (2.75,0) rectangle (3.15,1.50);
      \node[rotate=25,anchor=east] at (0.82,-0.16) {GPT-5.5};
      \node[rotate=25,anchor=east] at (1.92,-0.16) {Gemini};
      \node[rotate=25,anchor=east] at (3.02,-0.16) {Claude};
      \node at (0.75,0.70) {0.035};
      \node at (1.85,2.18) {0.135};
      \node at (2.95,1.66) {0.100};
    \end{scope}
  \end{tikzpicture}}
  \caption{Mechanism-probe visual summary. The three panels separate language removal, observation-noise robustness, and adversarial false-audit sensitivity.}
  \label{fig:mechanism_probe_summary}
\end{figure}
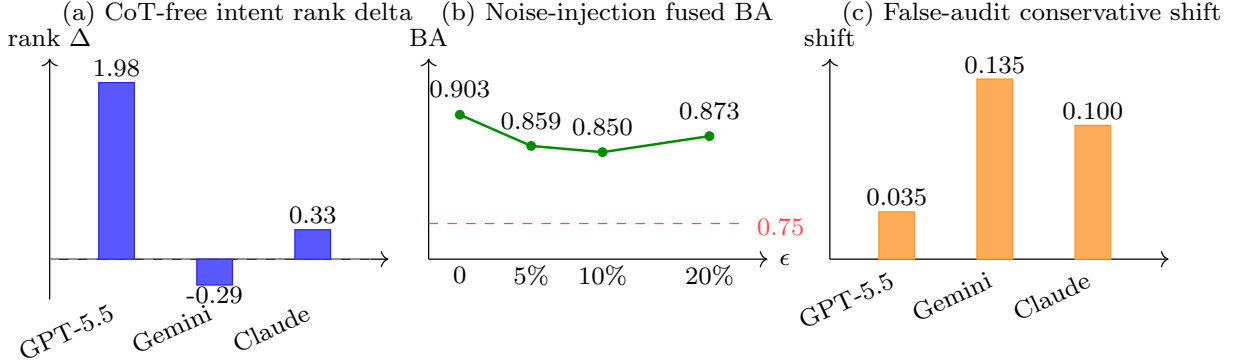

Figure~\ref{fig:mechanism_probe_summary} compresses the three most important mechanism probes into a visual check. The CoT-free panel shows why the paper avoids a single universal claim about intent geometry: GPT-5.5 retains a strong target-weight contraction signal, Claude retains a weaker one, and Gemini does not. The noise panel shows that the fused diagnostic stays above the 0.75 balanced-accuracy threshold even under 20\% Gaussian perturbation. The contrarian panel shows that all three models shift conservatively under false severe audits, exposing a trust-calibration attack surface.

\subsection{Representation-Manifold Signatures}

Table~\ref{tab:embedding_manifold} extends the centroid analysis with trajectory geometry. Two patterns are notable. First, several pre-drawdown windows show local acceleration: in Model A risk-aware, plan-step distance rises from 0.874 in normal states to 1.028 before the trough, a 1.176 ratio; in Model B no-risk, the plan ratio is 1.127. Second, pre-drawdown neighborhoods often have lower effective rank than normal neighborhoods, for example Model B no-risk plan rank falls from 26.79 to 17.90. This is consistent with a cognitive-narrowing interpretation: as the agent approaches failure, its planning manifold can become lower-dimensional, suggesting a shrinking set of decision modes rather than a smooth extrapolation of normal behavior. The analogy is tunnel vision in a high-pressure decision process. In the language of representation geometry, the agent appears to lose decision diversity before failure; this connects the financial setting to broader work on representation anisotropy and collapse while remaining an observable behavioral diagnostic rather than a claim about hidden activations.

The fused representation shows a similar but moderated pattern because structured risk and market features stabilize the text embedding. Model A risk-aware fused velocity ratio is 1.167, while Model B risk-aware fused ratio is close to one. This distinction supports the paper's broader claim: external audit features can make failure signatures more measurable and less dependent on free-form text alone. The nearest-neighbor phase-purity metric remains weak in the maximum-anchor view, which is useful negative evidence: pre-drawdown points are shifted and geometrically distorted, but they do not yet form a clean separate cluster.

\begin{table}[t]
  \centering
  \caption{Representation-manifold diagnostics around drawdowns. Step distance is local embedding velocity; effective rank estimates local dimensionality.}
  \label{tab:embedding_manifold}
  \resizebox{\linewidth}{!}{
  \begin{tabular}{llrrrrr}
    \toprule
    Case & View & Path length & Normal step & Pre step & Ratio & Normal/pre rank \\
    \midrule
    Model A risk-aware & Plan & 45.103 & 0.874 & 1.028 & 1.176 & 27.68 / 20.60 \\
    Model A risk-aware & Fused & 32.844 & 0.643 & 0.750 & 1.167 & 14.62 / 14.01 \\
    Model B risk-aware & Plan & 46.955 & 0.924 & 0.884 & 0.957 & 27.80 / 23.19 \\
    Model B risk-aware & Fused & 33.911 & 0.669 & 0.685 & 1.024 & 14.91 / 12.48 \\
    Model B no-risk & Plan & 50.834 & 0.985 & 1.110 & 1.127 & 26.79 / 17.90 \\
    Model B no-risk & Fused & 38.142 & 0.745 & 0.829 & 1.113 & 14.73 / 12.02 \\
    \bottomrule
  \end{tabular}
  }
\end{table}

\subsection{Audit and Reproducibility Coverage}

The trajectory collection contains 98 tracked raw runs and reports risk lifecycle coverage, trajectory reproducibility coverage, and agent trace coverage of 1.0 for all aggregate experiment families. This means that every evaluated step contains the required risk reports, reproducibility fields, and agent trace fields. These coverage metrics are intentionally strict. They turn auditability into a measured property of the benchmark rather than a narrative claim.

\subsection{Financial-Audit Agent Skill Challenge}

The public artifact also includes a separate skill-task suite that evaluates LLMs as financial-audit agents rather than as stock pickers. This experiment asks a different question from the trading matrix: can a model reliably inspect a trajectory, find risk-gate edits, recognize rejected or partial fills, respect the boundary between stress simulation and calibrated execution evidence, reproduce an artifact from commit/hash/command metadata, and weaken unsupported claims before they become leaderboard language? The tasks therefore use TradeArena skills as reviewer workflows. They are not injected into the trading-agent prompt, and using such a skill inside a benchmarked trading agent would have to be recorded as part of that agent's prompt or retrieval configuration.

The tracked public runs contain aggregate scores only; raw provider prompts and responses are treated as private debugging artifacts. The standard suite has 12 public artifact-review tasks, while the challenge suite has eight adversarial tasks that deliberately tempt a model into profitability overclaims, stress-model calibration overclaims, public-artifact privacy leaks, dirty reproduction claims, and market-rule overgeneralization. Table~\ref{tab:audit_skill_matrix} summarizes 500 Poe-mediated calls: 180 standard-suite calls, 240 challenge-suite calls, 72 appended follow-up calls, and 8 Claude adversarial follow-up calls. The standard run is broad and mostly diagnostic: Gemini 3.1 Pro averages 88.3\%, GPT-5.5 averages 85.0\%, and the remaining models expose more failures on claim, reproduction, and plugin tasks. The challenge suite is intentionally harder; scores compress into a 73.8--80.4\% band, and every model has at least one hard failure. The follow-up rows are useful because they show repeat sensitivity: Gemini and Kimi improve on the appended third sample, GPT-5.5 falls, and Claude Opus 4.7 scores 95.0\% on an additional adversarial-claim sample after showing high variance in the two-sample challenge run.

\begin{table}[t]
  \centering
  \caption{Poe-mediated financial-audit agent skill results. Scores measure artifact-review reliability, not trading profitability. The Claude follow-up row is a single appended adversarial-claim-boundary sample rather than a full three-variant repeat.}
  \label{tab:audit_skill_matrix}
  \scriptsize
  \begin{tabular}{lcccc}
    \toprule
    Model & Standard & Challenge & Follow-up & Hard fails \\
    \midrule
    GPT-5.5 & 85.0\% & 77.9\% & 72.5\% & 7 \\
    Gemini 3.1 Pro & 88.3\% & 80.4\% & 82.5\% & 6 \\
    Kimi K2.5 & 75.6\% & 73.8\% & 80.0\% & 8 \\
    GLM-5 & 79.4\% & 77.5\% & -- & 7 \\
    Claude Opus 4.7 & 77.2\% & 78.3\% & 95.0\% & 7 \\
    \bottomrule
  \end{tabular}
\end{table}

The skill results sharpen the paper's positioning. They show that frontier models can often identify mechanical audit evidence, especially risk edits and execution-boundary violations, yet remain fragile under adversarial wording about claims and reproduction. This is a reliability finding rather than a model-ranking claim. It says that TradeArena's trajectory schema can support a second benchmark axis: \emph{LLMs as auditors of financial-agent evidence}. That axis is valuable precisely because it penalizes fluent but unsupported interpretations, such as reading a stress-only execution row as a calibrated transaction-cost result or reading a cached provider row as a stable scientific claim about model trading ability.

\section{Positioning and Limitations}

The current version should be read as a research study of LLM financial decision dynamics, supported by an auditable experimental substrate, rather than as evidence of deployable trading performance. Its strongest claims concern representation signatures, risk-feedback alignment, and portfolio-reasoning failure modes. The main limitations are:
\begin{itemize}
  \item many ablation families use synthetic data so that component effects are controlled, although the synthetic suite now includes heterogeneous volatility and tail regimes;
  \item the historical experiment uses daily Yahoo Finance data with weekly decisions rather than intraday limit-order-book data;
  \item the intraday frontier LLM probes now cover a 40-hour week, but they still use hourly Yahoo Finance bars rather than limit-order-book data and currently cover two cached Poe-mediated models;
  \item the crisis-scene visual probes are diagnostic snapshots with 12 completed decisions per trajectory across GPT-5.5, Gemini 3.1 Pro, Claude Opus 4.7, and DeepSeek V4 Pro; they are included to visualize mechanism traces on real stress paths, not to establish long-horizon economic performance;
  \item direct-provider model-backed sanity checks use a short 52-week window, while the broader frontier matrix is routed through Poe rather than direct vendor APIs;
  \item the cached Poe-mediated frontier model matrix may include provider routing, wrapper prompts, truncation behavior, latency, billing availability, and model-version aliases that differ from direct vendor APIs; accordingly, the frontier conclusions are about the recorded Poe-mediated agent channel rather than hidden vendor internals, and any provider-side live-call failures are logged as missing cases rather than silently imputed;
  \item the BGE-M3 Transformer embedding validation now covers GPT-5.5 and Gemini 3.1 Pro but remains a subset cross-space agreement probe; scaling it to all LLM trajectories and additional dense embedding families such as \texttt{text-embedding-3-small} is a natural next validation target;
  \item the white-box hidden-state probe uses Qwen2.5-0.5B as an open surrogate over recorded decision texts; it does not expose the proprietary hidden activations of cached frontier models;
  \item the CoT-free, noise-injection, and contrarian-audit probes now include GPT-5.5, Gemini 3.1 Pro, and Claude Opus 4.7 over the 52-week historical task; broader provider-direct sweeps are needed before treating the effects as universal;
  \item the financial-audit skill challenge is a rubric-based artifact-review benchmark, not evidence that skills improve trading performance; if a benchmarked agent uses a skill as context, that context must be logged as part of the model configuration;
  \item the unsupported-claim proxy is audit-grounded and reproducible, but it should be interpreted as a closed-log factual-violation detector rather than a validated open-domain semantic hallucination benchmark unless the exported human annotation file is completed;
  \item comparisons to external trading-agent frameworks, deep RL policies, and realistic broker APIs remain future work;
  \item statistical uncertainty is reported for synthetic seed sweeps, rolling windows, and representation anchors, but the model-backed trading horizons remain short;
  \item results should be interpreted as benchmark diagnostics, not investment evidence.
\end{itemize}

These limitations are also the most direct path for strengthening the paper: add a FinRL-style PPO policy alongside the Markowitz baseline, complete the 50-step human hallucination calibration, expand the intraday probes to more direct-provider models, and test limit-order-book data. The modular analyst interface already separates provider access from the rest of the benchmark, so cached Poe-mediated rows, live Poe rows, DeepSeek rows, or direct vendor APIs can be swapped without changing the risk, execution, logging, or evaluation layers.

\section{Conclusion}

This paper argues that LLM trading agents should be studied through their decision dynamics, not only through their return curves. Across historical, synthetic, intraday, and crisis-scene experiments, including the 2022 Tech/Rates drawdown and 2023 SVB/regional-bank shock, the strongest evidence is not that LLMs trade profitably, but that their failures leave measurable traces: planning representations shift before drawdowns, rolling pre-failure anchors show effective-rank contraction across hash, LSA, BGE-M3 Transformer, and white-box hidden-state spaces, lexical controls rule out simple token-repetition collapse, CoT-free target weights sometimes preserve intent-space contraction, noisy OHLCV perturbations do not erase fused early-warning accuracy, structured risk feedback alters subsequent intent, placebo and contrarian feedback expose miscalibrated over-alignment, DeepSeek V4 Pro adds a direct-provider comparison to the Poe-mediated frontier matrix, high-dimensional portfolios reveal correlation-blind concentration relative to explicit risk control and a Markowitz covariance baseline, and financial-audit skill tasks expose whether frontier models can review trajectories without overclaiming the evidence. TradeArena is the substrate that makes these traces observable and replayable.

The broader alignment lesson is that constraint-following language is not the same as constraint-grounded reasoning. In our setting, benign placebo feedback can make an agent sound more risk-aware while adversarial false audits can push it toward unjustified conservatism; both are forms of an alignment tax in which textual compliance is purchased at the cost of distorted decision geometry. This connects financial trading agents to a wider problem in LLM safety: prompt-level rules, RLHF-style preferences, and constitutional instructions may shape the surface vocabulary of alignment without guaranteeing that the model preserves a coherent causal link between evidence, constraint, and action.

Future work will test whether the same representation signatures hold under longer horizons, direct model-provider APIs, richer market data, larger proprietary and open-source embedding models, deep RL baselines, and external agent frameworks.


\begin{thebibliography}{99}

\bibitem{markowitz1952}
H. Markowitz.
Portfolio selection.
\emph{The Journal of Finance}, 7(1):77--91, 1952.

\bibitem{sharpe1966}
W. F. Sharpe.
Mutual fund performance.
\emph{The Journal of Business}, 39(1):119--138, 1966.

\bibitem{kahneman1979}
D. Kahneman and A. Tversky.
Prospect theory: an analysis of decision under risk.
\emph{Econometrica}, 47(2):263--292, 1979.

\bibitem{almgren2001}
R. Almgren and N. Chriss.
Optimal execution of portfolio transactions.
\emph{Journal of Risk}, 3(2):5--39, 2001.
doi:10.21314/JOR.2001.041.

\bibitem{bailey2014}
D. H. Bailey, J. M. Borwein, M. Lopez de Prado, and Q. J. Zhu.
The probability of backtest overfitting.
\emph{Journal of Computational Finance}, 20(4):39--69, 2017.
doi:10.21314/JCF.2016.322.

\bibitem{react2023}
S. Yao, J. Zhao, D. Yu, N. Du, I. Shafran, K. Narasimhan, and Y. Cao.
ReAct: Synergizing reasoning and acting in language models.
\emph{International Conference on Learning Representations}, 2023.

\bibitem{toolformer2023}
T. Schick, J. Dwivedi-Yu, R. Dess\`i, R. Raileanu, M. Lomeli, E. Hambro, L. Zettlemoyer, N. Cancedda, and T. Scialom.
Toolformer: Language models can teach themselves to use tools.
\emph{Advances in Neural Information Processing Systems}, 2023.

\bibitem{generativeagents2023}
J. S. Park, J. O'Brien, C. J. Cai, M. R. Morris, P. Liang, and M. S. Bernstein.
Generative agents: Interactive simulacra of human behavior.
\emph{ACM Symposium on User Interface Software and Technology}, 2023.

\bibitem{ethayarajh2019}
K. Ethayarajh.
How contextual are contextualized word representations? Comparing the geometry of BERT, ELMo, and GPT-2 embeddings.
\emph{Proceedings of EMNLP-IJCNLP}, pages 55--65, 2019.

\bibitem{papyan2020}
V. Papyan, X. Y. Han, and D. L. Donoho.
Prevalence of neural collapse during the terminal phase of deep learning training.
\emph{Proceedings of the National Academy of Sciences}, 117(40):24652--24663, 2020.

\bibitem{finrl2020}
X.-Y. Liu, H. Yang, Q. Chen, R. Zhang, L. Yang, B. Xiao, and C. D. Wang.
FinRL: A deep reinforcement learning library for automated stock trading in quantitative finance.
\emph{NeurIPS Workshop on Deep Reinforcement Learning}, 2020.

\bibitem{finrlmeta2022}
X.-Y. Liu, Z. Xia, J. Rui, J. Gao, H. Yang, M. Zhu, C. D. Wang, Z. Wang, and J. Guo.
FinRL-Meta: Market environments and benchmarks for data-driven financial reinforcement learning.
\emph{Advances in Neural Information Processing Systems Datasets and Benchmarks}, 2022.

\bibitem{qlib2020}
X. Yang, W. Liu, D. Zhou, J. Bian, and T.-Y. Liu.
Qlib: An AI-oriented quantitative investment platform.
\emph{arXiv preprint arXiv:2009.11189}, 2020.

\bibitem{fingpt2023}
H. Yang, X.-Y. Liu, and C. D. Wang.
FinGPT: Open-source financial large language models.
\emph{arXiv preprint arXiv:2306.06031}, 2023.

\bibitem{tradingagents2024}
Y. Xiao, E. Sun, D. Luo, and W. Wang.
TradingAgents: Multi-Agents LLM Financial Trading Framework.
\emph{arXiv preprint arXiv:2412.20138}, 2024.

\bibitem{bge_m3}
J. Chen, S. Xiao, P. Zhang, K. Luo, D. Lian, and Z. Liu.
M3-Embedding: Multi-Linguality, Multi-Functionality, Multi-Granularity Text Embeddings Through Self-Knowledge Distillation.
\emph{arXiv preprint arXiv:2402.03216}, 2024.
Model card: \url{https://huggingface.co/BAAI/bge-m3}. Accessed May 17, 2026.

\bibitem{qwen25}
A. Yang et al.
Qwen2.5 technical report.
\emph{arXiv preprint arXiv:2412.15115}, 2024.
Model card: \url{https://huggingface.co/Qwen/Qwen2.5-0.5B-Instruct}. Accessed May 17, 2026.

\bibitem{yfinance_docs}
R. Aroussi.
yfinance documentation.
\url{https://ranaroussi.github.io/yfinance/}. Accessed May 17, 2026.

\end{thebibliography}
\end{document}